\documentclass[lettersize,journal]{IEEEtran}
\usepackage{array}
\usepackage[caption=false,font=normalsize,labelfont=sf,textfont=sf]{subfig}
\usepackage{textcomp}
\usepackage{stfloats}
\usepackage{url}
\usepackage{verbatim}
\usepackage{graphicx}
\usepackage{float}
\usepackage{ctable}
\usepackage{wrapfig}
\usepackage{amssymb}
\usepackage{multirow}
\usepackage{amsfonts}
\usepackage{amsmath}
\usepackage{booktabs}
\usepackage{makecell}
\usepackage{array}
\usepackage{amssymb}
\usepackage{amsthm}
\usepackage{booktabs}
\usepackage{algorithm}
\usepackage[noend]{algpseudocode}

\graphicspath{{Figs/}}
\newcommand{\vect}[1]{\boldsymbol{#1}}
\def\BibTeX{{\rm B\kern-.05em{\sc i\kern-.025em b}\kern-.08em
    T\kern-.1667em\lower.7ex\hbox{E}\kern-.125emX}}
\usepackage{balance}
\begin{document}
\title{MUSE-Net: Missingness-aware mUlti-branching Self-attention Encoder for Irregular Longitudinal Electronic Health Records}


\author{Zekai Wang}
\author{Tieming Liu}
\author{Zekai Wang, Tieming Liu, Bing Yao
\thanks{Corresponding author: byao3@utk.edu; \\Zekai Wang is with the Charles F. Dolan School of Business, Fairfield University.\\Bing Yao are with the Department of Industrial \& Systems Engineering, The University of Tennessee, Knoxville, TN, 37996 USA.\\Tieming Liu is with the School of Industrial Engineering and Management, Oklahoma State University, Stillwater, OK 74078.}}

\maketitle

\begin{abstract}
The era of big data has made vast amounts of clinical data readily available, particularly in the form of electronic health records (EHRs), which provides unprecedented opportunities for developing data-driven diagnostic tools to enhance clinical decision making. However, the application of EHRs in data-driven modeling faces challenges such as irregularly spaced multi-variate time series, issues of incompleteness, and data imbalance. Realizing the full data potential of EHRs hinges on the development of advanced analytical models. In this paper, we propose a novel \(\mathbf{M}\)issingness-aware m\(\mathbf{U}\)lti-branching \(\mathbf{S}\)elf-Attention \(\mathbf{E}\)ncoder (MUSE-Net) to cope with the challenges in modeling longitudinal EHRs for data-driven disease prediction. The proposed MUSE-Net is composed by four novel modules including: (1) a multi-task Gaussian process (MGP) with missing value masks for data imputation; (2) a multi-branching architecture to address the data imbalance problem; (3) a time-aware self-attention encoder to account for the irregularly spaced time interval in longitudinal EHRs; (4) interpretable multi-head attention mechanism that provides insights into the importance of different time points in disease prediction, allowing clinicians to trace model decisions. We evaluate the proposed MUSE-Net using both synthetic and real-world datasets. Experimental results show that our MUSE-Net outperforms existing methods  that are widely used to investigate longitudinal signals.
\end{abstract}

\def\abstractname{Note to Practitioners}
\begin{abstract}
This article is motivated by the growing need for robust machine learning models capable of handling the complexities of real-world EHRs, including irregular time intervals, missing data, and class imbalance. The proposed MUSE-Net model integrates advanced imputation via multi-task Gaussian processes with missingness masks, a time-aware self-attention encoder, and a multi-branching framework to enhance predictive accuracy and robustness. Additionally, MUSE-Net leverages an interpretable multi-head attention mechanism to provide transparent decision-making, allowing clinicians to trace model predictions back to key time points. This framework offers a practical and trustworthy solution for data-driven disease prediction and clinical decision support.
\end{abstract}


\section{Introduction}
Rapid advancements in sensing and information technology have ushered us into an era of data explosion where a large amount of data is now easily available and accessible in the clinical environment \cite{yang2023sensing,yao2021constrained,yao2020spatiotemporal}. The wealth of healthcare data offers new avenues for developing data-driven methods for automated disease diagnosis. For instance, there have been growing research interests in harnessing electronic health records (EHRs) to create data-driven solutions for clinical decision support \cite{rajkomar2018scalable} in detecting heart disease \cite{wang2023hierarchical}, sepsis \cite{wang2021multi}, and diabetes \cite{wang2024multi}. EHRs serve as digital repositories of a patient's medical information including demographics, medications, vital signs, and lab results \cite{yadav2018mining,shickel2017deep,yao2017characterizing}, curated over time by healthcare providers, leading to a longitudinal database. With rich information about a patient's health trajectory, longitudinal EHRs present unique opportunities to analyze and decipher clinical events and patterns within large populations through data-driven machine learning.

However, data mining of longitudinal EHRs poses distinct challenges due to the observational nature of EHRs. Unlike well-defined, randomized experiments in clinical trials that collect data on a fixed schedule and ensure high data quality, EHRs are recorded only when patients receive care or doctors provide services. The information collected and the timing of its collection are not determined by researchers, resulting in EHRs that are highly heterogeneous \cite{xiao2018opportunities} and further introducing the following challenges in  data-driven decision-making:

(1) \textbf{Irregularly spaced time series}.
EHR data are often documented during irregular patient visits, leading to non-uniform time intervals between successive measurements and a lack of synchronization across various medical variables or among different patients. Traditional time series models face challenges when applied to irregular longitudinal data because they typically assume a parametric form of the temporal variables, making them difficult to effectively account for highly heterogeneous and irregular time intervals across different variables. Additionally, the widely used deep learning models such as convolution neural networks (CNNs) and recurrent neural networks (RNNs) for mining sequential or time series data are designed by assuming consecutive data points are collected at a uniform time interval. Those deep learning models do not consider the elapsed time between records and are less effective in modeling irregular longitudinal EHRs.

(2) \textbf{Incomplete data and imbalanced class distributions}. EHRs suffer from the issues of missing values and imbalanced data.  Due to the nature of clinical practice, not all information is recorded for every patient visit, leading to incomplete datasets. Additionally, EHR data often exhibit a significantly imbalanced distribution, with certain health outcomes or characteristics being underrepresented. For instance, rare diseases or adverse drug reactions \cite{schieppati2008rare, wang2023reslife} may have very few instances compared to more prevalent conditions. In the existing literature, a wide array of statistical and machine learning techniques have been designed to tackle the missing value and imbalanced data issues \cite{emmanuel2021survey,he2009learning}, which, however, are less applicable in the context of modeling irregular longitudinal EHRs. The presence of missing values and imbalanced class distributions will introduce further difficulties in effective model training using longitudinal EHRs, leading to biased or inaccurate predictions if not properly addressed.

To address the challenges presented by longitudinal EHRs, this paper introduces a novel framework of \(\mathbf{M}\)issingness-aware m\(\mathbf{U}\)lti-branching \(\mathbf{S}\)elf-Attention \(\mathbf{E}\)ncoder (MUSE-Net) for data-driven disease prediction. First, multi-task Gaussian processes (MGPs) are employed for missing value imputation in irregularly sampled time series. Second, we propose to add missing value masks that record the locations of missing observations as another input stream to our predictive model, which enables the learning of correlations between non-missing and missing values for mitigating the impact of possible imputation errors incurred from the imputation procedure on the prediction performance. Third, we propose to integrate a time-aware self-attention encoder with a multi-branching classifier to address the imbalanced data issue and further classify the irregular longitudinal EHRs for disease prediction. Furthermore, MUSE-Net employs an interpretable multi-head attention mechanism \cite{lim2021temporal} that highlights critical time points in disease prediction, offering transparency in decision-making and enabling clinicians to trace model outputs back to influential time points. This interpretability fosters trust and facilitates the integration of MUSE-Net into real-world clinical applications.
We evaluate our proposed framework using both simulation data and real-world EHRs. Experimental results show that our proposed method significantly outperforms existing approaches that are widely used in current practice.

\section{Research Background} 

The integration of data-driven modeling and EHRs has transformed the healthcare field, which provides unprecedented opportunities for clinical decision support \cite{yadav2018mining, landi2020deep, rajpurkar2022ai}. Extensive research has been conducted to develop data-driven models using non-longitudinal EHRs that consist of static or cross-sectional health information \cite{banda2018advances,wang2022multi}. For example, Huang et al. \cite{huang2007feature} employed naive Bayes, decision tree, and nearest neighbor algorithms incorporating feature selection methods to determine key factors affecting type II diabetes control and identify individuals who exhibit suboptimal diabetes control status. Hong et al. \cite{hong2019developing} developed a multi-class classification method to analyze clinical data in identifying patients with obesity and various comorbidities using logistic regression, support vector machine, and decision tree. A comprehensive review on machine learning of non-longitudinal EHRs can be referred to \cite{xiao2018opportunities, rajkomar2019machine}. However, those methods are limited in capturing temporal patterns in health status over time. This limitation can result in less accurate predictions for conditions that are heavily dependent on longitudinal health trajectories \cite{banda2018advances, das2022critical}. Moreover, traditional machine learning approaches often require manual feature engineering, which is time-consuming and prone to error. 

With the growing availability of longitudinal EHRs, increasing interests have been devoted to developing advanced models to capture temporal information for disease prediction. Owing to the strong capability in pattern recognition, deep learning has been widely explored to mine complexly structured data \cite{xie2022physics,xie2023automated,xie2022physics2,wang2023hierarchical}. Advanced network architectures have been crafted for modeling time series or sequential data. For example, RNNs including long short-term memory (LSTM) are among the most commonly used models to analyze medical time series for various clinical tasks \cite{choi2017using, che2018recurrent, chen2022prediction}. Additionally, temporal convolutional networks (TCNs) have been recognized as a robust alternative to RNNs for modeling longitudinal signals \cite{wang2024multi,wang2021multi,rosnati2021mgp}. 
However, traditional RNNs and TCNs are designed with the assumption that the records are collected at a constant rate and require the neighboring samples to appear at fixed distances to facilitate the convolution or recurrent operations. This assumption is not valid in many real-world databases, making traditional RNNs or TCNs less effective in modeling irregular-spaced longitudinal EHRs.

To cope with the issue of irregular time intervals, many modified RNN architectures have been developed. For example, the time-aware LSTM (TLSTM) was designed to account for non-uniform sampling intervals through a time decay mechanism \cite{baytas2017patient}. Che et al. \cite{che2018recurrent} also implemented a decay mechanism and proposed a GRU-D model, allowing the network to better capture temporal dependencies even when data points were incomplete or irregularly sampled. A comprehensive review of modified RNN architectures for irregular sampled time series can be found in  \cite{sun2020review, weerakody2021review}. However, most existing time-aware RNNs assume that the impact of historical risk factors on disease prediction proportionally decays over time, which may not be true in describing complex disease trajectories. Additionally, RNNs have been widely recognized as computationally inefficient for modeling large-scale, long sequential data due to their sequential processing nature. 

The self-attention encoder, a cornerstone of the transformer architecture \cite{vaswani2017attention}, has revolutionized the field of natural language processing (NLP). Unlike RNNs and TCNs that process data sequentially and require uniform sampling intervals, the self-attention mechanism allows the model to weigh the importance of different parts of the input sequences relative to each other. This feature enables parallel processing and enhances the model's ability to capture long-range dependencies in irregular longitudinal signals. For instance, Li \textit{et al.} proposed to transform time series data into images and adopted the Vision Transformer to model irregularly sampled time series signals \cite{li2024time}. Tipirneni and Reddy developed a method that integrated a continuous value embedding technique with self-attention to model irregularly sampled clinical time series by  treating them as a set of observation triplets (time, variable, and value) \cite{tipirneni2022self}. Huang et al. developed a Deformable Neighborhood Attention Transformer to capture local and global dependencies in medical time series data through deformable attention mechanisms \cite{huang2024dna}. Manzini \textit{et al.} introduced the Diabetic Attention with Relative representation Encoder for Type 2 Diabetes patients \cite{manzini2025deep}. Those self-attention models demonstrate significant improvements over traditional methods, particularly in handling irregular longitudinal datasets. 


Despite the strengths of self-attention encoders, most deep learning models assume well-structured data, making them less effective for longitudinal EHRs that are often incomplete, imbalanced, and irregularly sampled. Existing methods rarely integrate strategies to handle missing values, imbalanced class distributions, and temporal irregularity simultaneously, limiting their reliability in clinical applications. There remains a critical need for a framework that not only addresses these challenges holistically but also ensures robust and interpretable decision-making for real-world healthcare prediction tasks.

\begin{figure*}[!ht]
	\begin{center}
		\includegraphics[width=5.5in]{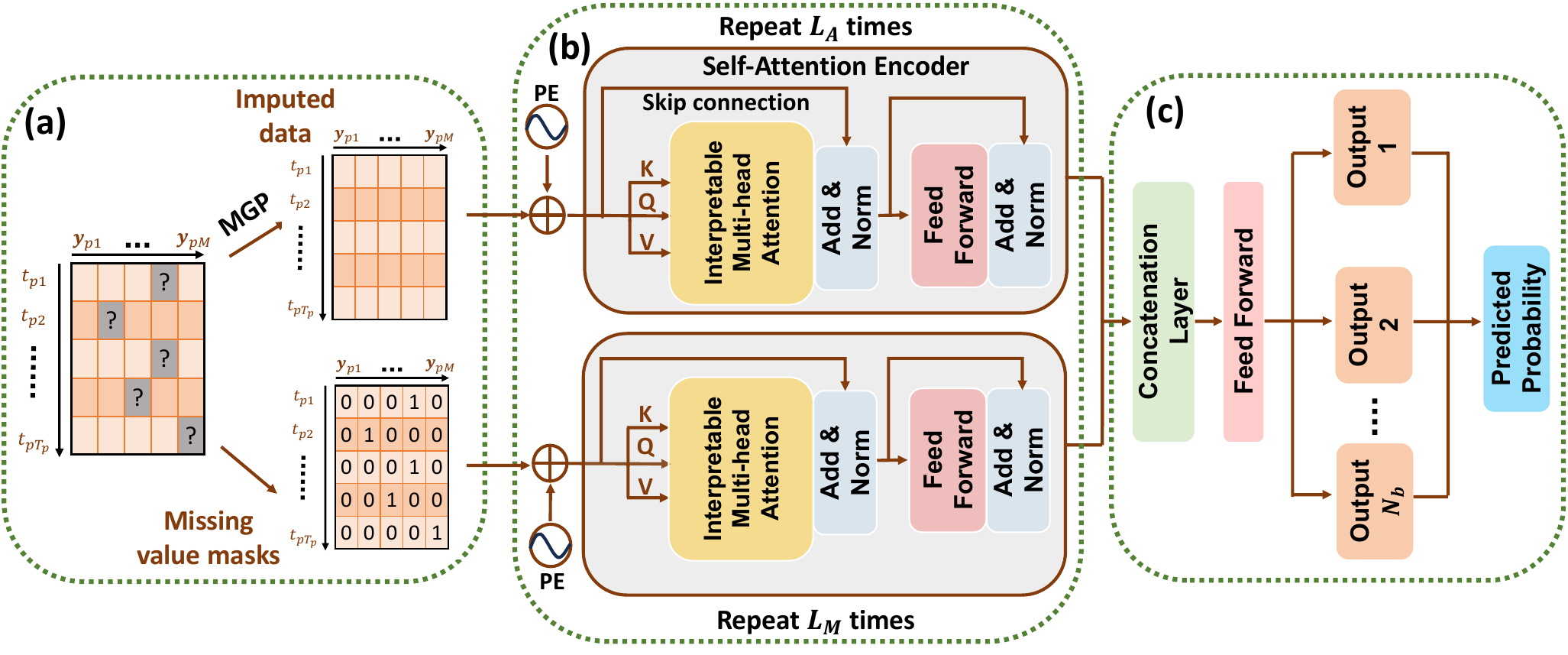}
		\caption{(a) MGP for data imputation and missing value masks generation; The imputed data and missing value masks are processed by MUSE-Net, which consist of (b) Missingness-aware Self-attention Encoder with Interpretable Multi-head Attention and (c) Multi-branching output layer.}
		\label{Fig:Model_Architecture}
        \vspace{-0.5cm}
        
	\end{center}    
\end{figure*}

\section{Research Methodology} 
Suppose that there are \( N \) patients indexed by \( p \in \{1, \ldots, N\} \) and we denote the dataset as \( \mathcal{D} = \{ (\vect{t}_p, Y_p,  l_p) \}_{p=1}^{N} \). Each patient is associated with longitudinal data described by the tuple \( (\vect{t}_p, Y_p,  l_p) \): \(\vect{t}_p = [t_{p,1}, t_{p,2}, \cdots, t_{p,T_p}]\) is the time index set for patient \(p\); \(Y_p = [\vect{y}_{p,1}, \vect{y}_{p,2}, \cdots, \vect{y}_{p,M}] \in \mathbb{R}^{T_p \times M}\) represents the values of \(M\) medical variables at \(\vect{t}_p\); \( l_p \) is the binary label with \( l_p = 1 \) indicating patient \( p \) is positive (e.g., with disease) and otherwise \( l_p = 0 \). The length of the time series for each patient is highly variable (i.e., \(T_p \neq T_{p^{'}}, ~\text{if}~p\neq p^{'}\) ), and the times series are often irregularly spaced (i.e., \(t_{p, i+1} - t_{p,i} \neq t_{p, j+1} - t_{p,j}, ~\text{if}~j \neq i\)) in real-world longitudinal EHR databases. Additionally, the EHR dataset is often incomplete, and we denote the complete set of values of the $M$ variables as \(\vect{y}_p = \text{Vect}(Y_p) =[\vect{y}_{p,1}^T, \vect{y}_{p,2}^T,  \cdots, \vect{y}_{p,M}^T]^T \in \mathbb{R}^{T_pM}\). Then, we define an index set that contains observed values as \(I_o = \{(t,m) | \text{ if } [Y_p]_{t,m} \text{ is observed}\}\) and the corresponding values are denoted by \(\vect{y}_p^o \in \mathbb{R}^{O_p}\). Similarly, we define \(I_u = \{(t,m) |\text{ if } [Y_p]_{t,m} \text{ is missing}\}\)  as the index set of missing values, and the missing value vector is denoted by \(\vect{y}_p^u \in \mathbb{R}^{U_p}\), where \(U_p + O_p = T_pM\). Fig. \ref{Fig:Model_Architecture} shows the flowchart of the proposed MUSE-Net to classify irregularly spaced and incomplete longitudinal data. Each component in the flowchart is described in detail in the following subsections.

\subsection{MGP for Missing Value Imputation}
A Gaussian process (GP) is a flexible non-parametric Bayesian model where any collection of random variables follows a joint Gaussian distribution \cite{williams2006gaussian,xie2023hierarchical,yao2021spatiotemporal}, which have been widely used to model complex time series data \cite{moor2019early, zhang2022multi}. GP-based temporal models provide a way to determine the distribution of a variable at any arbitrary point in time, making them intrinsically capable of dealing with missing value imputations that involve irregularly spaced longitudinal data.  Note that single-task GPs are limited in their ability to model correlations across multiple related tasks (i.e., different medical variables), making them less effective for modeling multi-variate EHR data. To account for the multi-variate nature, we adopt a multi-task GP (MGP) \cite{bonilla2007multi} to capture both variable interactions and temporal correlations for imputing missing values in irregular multi-variate longitudinal EHRs.

We denote \(f_{pm}(t)\) as a latent function representing the true values of variable \(m\) for patient \(p\) at time \(t\), and a patient-independent MGP prior is placed over the latent function \(f_{pm}(t)\) with a zero mean and the covariance function as:
\begin{eqnarray}
\text{cov}(f_{pm}(t), f_{pm^{'}}(t^{'})) &=& \mathcal{K}_M(m, m^{'})\mathcal{K}_t(t, t^{'})  \nonumber \\ 
y_{pm}(t) &\sim& \mathcal{N}(f_{pm}(t), \sigma^{2}_{m})
\label{MGP Prior}
\end{eqnarray}
where \(y_{pm}(t)\) is the observed value of variable \(m\) for patient \(p\) at time \(t\), \(\sigma^2_m\) is the noise term for the \(m\)th task (variable), \(\mathcal{K}_M(m, m^{'})\) captures the similarities between tasks, and \(\mathcal{K}_t(t, t^{'})\) is a temporal correlation function, which is defined as 
\(\mathcal{K}_t(t, t^{'}) = \text{exp}(-\frac{(t-t^{'})^2}{2\theta^2})\) with a lengthscale parameter \(\theta\). Hence, the prior distribution for the fully observed multivariate longitudinal records, \(\vect{y}^o_p \),  can be represented by:
\begin{eqnarray}
\vect{y}_p^o \sim \mathcal{N}(\vect{0}, \Sigma_p^o), ~~~~\Sigma_p^o =\mathcal{K}_M \odot \mathcal{K}_{O_p}  + E
\label{MGP Prior full}
\end{eqnarray}
We formulate the kernel function \(\Sigma_p^o\) for the observed records as the combination of the Hadamard product (i.e., element-wise multiplication) of \(\mathcal{K}_M\) and \(\mathcal{K}_{O_p}\), and a noise term $E$: \(\mathcal{K}_M\) is a \(O_p \times O_p\) positive semi-definite covariance matrix over medical variables, which is parameterized by a low-rank matrix \(B\) as \(LBB^TL^T\) where \(B \in \mathbb{R}^{M \times q}\), and \(L \in \mathbb{R}^{O_p \times M}\) is an indicator matrix with \(L_{im} = 1\) if the \(i^{\text{th}}\) observation belongs to task \(m\) and \(\sum_{m = 1}^{M} L_{im} = 1\); \(\mathcal{K}_{O_p}\)  is a \(O_p \times O_p\) squared exponential correlation matrix over time with elements defined as \(\mathcal{K}_{O_p}((m,t),(m',t'))=\mathcal{K}_t(t, t^{'})\) if $(m,t)\in I_o$ and $(m',t')\in I_o$; \(E\) is a noise matrix with \(L\vect{\sigma}^2\) on its main diagonal (\(\vect{\sigma}^2\ \in \mathbb{R}^{M}\) and \([\vect{\sigma}^2]_m=\sigma_m^2\) ). 

In the inference step, we impute the missing value for patient \(p\) using the posterior mean, \(\mu_{\vect{y}_p^u}\):
\begin{eqnarray}
\mu_{\vect{y}_p^u} &=& (\mathcal{K}_{M_{*}M} \odot \mathcal{K}_{U_pO_p})(\Sigma_p^o)^{-1}\vect{y}_p^o 
\label{MGP Posterior full}
\end{eqnarray}
where 
\(\mathcal{K}_{M_{*}M} = L_{*}BB^TL^T \in \mathbb{R}^{U_p \times O_p} \) is the task covariance matrix for missing values, and \(L_{*} \in \mathbb{R}^{U_p \times M}, L_{*,im} = 1\) if the \(i^{\text{th}}\) missing value belongs to task \(m\);  \(\mathcal{K}_{U_pO_p}\) represents the correlation matrix between the time points of the observed and missing values. The hyperparameters in the MGP model include the lower triangular matrix \(B\), error term \(\text{Diag}(E)\), and the lengthscale \(\theta\) in $\mathcal{K}_{O_p}$: {\small\(\vect{\Theta} = \{\text{Vect}(B), \text{Diag}(E), \theta \} \)}, which are learned by minimizing the negative log marginal likelihood given by (with derivative)
{\footnotesize
\begin{eqnarray}
\mathcal{L}(\vect{\Theta}) &=& \log |\Sigma_p^o| + (\vect{y}_p^o)^T (\Sigma_p^o)^{-1} \vect{y}_p^o + \frac{O_p}{2}\log(2\pi) \nonumber \\
\frac{d\mathcal{L}}{d\vect{\Theta} } &=& (\vect{y}_p^o)^T (\Sigma_p^o)^{-1} \frac{d\Sigma_p^o}{d\vect{\Theta}} (\Sigma_p^o)^{-1} \vect{y}_p^o + \text{Tr} \left( (\Sigma_p^o)^{-1} \frac{d\Sigma_p^o}{d\vect{\Theta}} \right)
\label{Training and gradient}
\end{eqnarray}}
Finally, the missing values in $Y_p$ are estimated as $\hat{\vect{y}}_p^u=\mu_{\vect{y}_p^u}$, which are combined with the observed values $\vect{y}_p^o$ to form an imputed dataset {\small\( \hat{\mathcal{D}} = \{ (\vect{t}_p, \hat{Y}_p,  l_p) \}_{p=1}^{N} = \{ ((\vect{t}_p^o, \text{Vect}^{-1}(\vect{y}_p^o)), (\vect{t}_p^u, \text{Vect}^{-1}(\hat{\vect{y}}_p^u)), l_p) \}_{p=1}^{N}\)}, where $\text{Vect}^{-1}(\cdot)$ is the inverse operation of $\text{Vect}(\cdot)$. To accelerate MGP Imputation, we adopt the black box matrix-matrix multiplication framework \cite{gardner2018gpytorch} to integrate the modified preconditioned conjugate gradient (mPCG) with GPU acceleration into the data imputation workflow. Please refer to Appendix \ref{appendix mgp} for the detailed procedure for GPU acceleration of MGP imputation.

\subsection{MUSE-Net}
 We further propose a MUSE-Net to process the imputed longitudinal EHRs. This model is designed to not only effectively capture crucial temporal correlations in multi-variate irregular longitudinal data but also recognize missingness patterns and account for the imbalanced data issue for enhanced classification performance. As shown in Fig. \ref{Fig:Model_Architecture}(b) and (c), this model consists of 3 key modules: time-aware self-attention encoder, missing value masks, and multi-branching outputs.

\subsubsection{Time-aware Self-attention Encoder}
  We propose to adapt the traditional self-attention encoder \cite{vaswani2017attention} to incorporate the information of elapsed time between consecutive records in irregular longitudinal EHRs, including three building blocks: 
  
  \textbf{Elapsed time-based positional encoding:}  The time order of the longitudinal sequence plays a crucial role in time series analysis. However, this information is often ignored in traditional attention-based encoders because it does not incorporate any recurrent or convolution operations. 
  To address this issue, an elapsed time-based positional encoding is added into the input sequence at the beginning of the self-attention encoder as shown in Fig. \ref{Fig:Model_Architecture}(b): 
    \begin{eqnarray}
PE_{t_{p,i},2m} = \text{sin}(t_{p,i}/10000^{2m/M}) \\ \nonumber
PE_{t_{p,i},2m + 1} = \text{cos}(t_{p,i}/10000^{2m/M})
   \label{pe}
\end{eqnarray}
where \(t_{p,i}\) is the time position of each observation to account for the irregular time interval, and \(2m\) and \(2m+1\) are the (\(2m\))-th and (\(2m+1\))-th variable dimensions, respectively. The sinusoid prevents positional encodings from becoming too large, introducing extra difficulties in network optimization. The embedded elapsed time-based positional features will then be combined with the imputed signals to generate the input of the self-attention module: $X_p=PE_p+\hat{Y}_p$, where $PE_p\in \mathbb{R}^{T_p\times M}$ is a matrix with elements defined in Eq. (\ref{pe}).

\textbf{Intepretable multi-head attention}: The self-attention is a mechanism to allow the network to learn dependencies in the longitudinal data. It maps a query set and a set of key-value pairs to an output. In a single-head self-attention, the key, query, and value matrices, denoted as \(K\in \mathbb{R}^{T_p\times M}\), \(Q\in \mathbb{R}^{T_p\times M}\), and \(V\in \mathbb{R}^{T_p\times M}\), are computed by taking input \(X_p \): 
    \begin{eqnarray}
        K=\mathcal{F}_K(X_p), ~~ Q=\mathcal{F}_Q(X_p),~~ V=\mathcal{F}_V(X_p)
    \end{eqnarray} 
   where $\mathcal{F}_K(\cdot)$, $\mathcal{F}_Q(\cdot)$, and $\mathcal{F}_V(\cdot)$ represent the network operations to calculate the key, query, and value matrices respectively, which are often selected as linear transformations. The corresponding output, \(\text{Output}_S \in \mathbb{R}^{T_p \times M}\), is computed by applying the scaled-dot production attention:
\begin{eqnarray}
\text{Output}_S = \text{softmax}(\frac{QK^T}{\sqrt{M}}) \times V
\label{KQV out}
\end{eqnarray}
  where $\text{softmax}(\vect{x})_i=\frac{e^{x_i}}{\sum_j e^{x_j}}$ outputs the weight assigned to each element in $V$.  The dot product in the softmax is scaled down by \(\sqrt{M}\). This is essential to prevent the dot product values from growing too large, especially when the dimensionality of \(\hat{Y}_p\) or $X_p$ is large, introducing instabilities in the training process. The self-attention mechanism enables the network to access all the information of the input with the flexibility of focusing on certain important elements over the entire sequence.
    
The multi-head attention module is an extension of single-head self-attention to capture different relations among multivariate time series. Specifically, multiple matrices for the keys, queries, and values are defined by applying multiple network operations, $\mathcal{F}_K^l(\cdot)$'s, $\mathcal{F}_Q^l(\cdot)$'s, and $\mathcal{F}_V^l(\cdot)$'s, to the input:
       \begin{eqnarray}
        K_l=\mathcal{F}_K^l(X_p)\in\mathbb{R}^{T_p\times \frac{M}{h}},~~   Q_l=\mathcal{F}_Q^l(X_p)\in\mathbb{R}^{T_p\times \frac{M}{h}} \\ \nonumber V_l=\mathcal{F}_V^l(X_p)\in\mathbb{R}^{T_p\times \frac{M}{h}}
    \end{eqnarray}
    where \(h\) is the number of attention heads and \(l \in \{1,\dots,h\}\). The output of each attention head is calculated by:
\begin{eqnarray}
\text{Output}_l = \text{softmax}(\frac{Q_lK_l^T}{\sqrt{M/h}}) \times V_l \in \mathbb{R}^{T_p \times \frac{M}{h}}
\label{KQV multi-out}
\end{eqnarray}
The final output of the multi-head attention, \(\text{Output}_M \in \mathbb{R}^{T_p \times M}\), is the multiplication of the concatenation of the outputs of all heads and a weight matrix, \(W^O \in \mathbb{R}^{M \times M}\):
\begin{eqnarray}
\text{Output}_M = [\text{Output}_1, \text{Output}_2, \cdots, \text{Output}_h]W^O
\label{KQV final multi-out}
\end{eqnarray}
This multi-head attention module enables the network to jointly attend to information from different subspaces of multivariate time series at different time points. 

However, traditional multi-head attention suffers from three limitations: (1) Lack of interpretability: the concatenation of head outputs makes it difficult to analyze how each head contributes to the final decision; (2) Independent value matrices: each head has its own $V_l$, leading to inconsistent feature extraction across heads; (3) Extra computation overhead: maintaining separate $V_l$'s increases parameter complexity. To address these issues, we adopt the Interpretable Multi-Head Attention. Instead of maintaining separate value matrices $V_l$,  Interpretable Multi-head Attention introduces a shared value matrix $V$ for all heads $V= X_pW^V$, where $W^V$ is a single weight matrix applied to the entire input. Additionally, 
the weighted key-query matrices across all heads are summed as:
\begin{eqnarray}
S = \sum_{l=1}^{h}\text{softmax}(\frac{Q_lK_l^T}{\sqrt{M/h}})
\label{sum key-query}
\end{eqnarray}
Then, the final output is computed by multiplying the aggregated attention score with the shared value matrix $V$:
\begin{eqnarray}
\text{Output}_M  = SVW^O
\label{intepretable output}
\end{eqnarray}

This modification improves interpretability by allowing a global attention map $S$ to capture how different query-key interactions influence the shared value matrix. Instead of treating each head’s output separately, interpretable multi-head attention integrates them in a structured and meaningful way.

\textbf{Feed-forward network}: To stabilize the training process, $\text{Output}_M$ will pass through a layer normalization operation \cite{ba2016layer}, and the normalized output will be combined with the original input using skip-connection \cite{he2016deep} to generate the output of the ``Add \& Norm" block (see Fig. \ref{Fig:Model_Architecture}(b)):
    \begin{eqnarray}
        \text{Output}_{AN} =\text{LayerNorm}(\text{Output}_M)+X_p
    \end{eqnarray}
  $\text{Output}_{AN}$ will serve as the input of a feed-forward network with GELU \cite{hendrycks2016gaussian} activation and additional ``Add \& Norm" blocks to further induce nonlinearity degree into the self-attention encoder. The multi-head attention and feed-forward modules will be repeated multiple times to improve the generalizability of the model for capturing more informative features of the input sequence.

\subsubsection{Missing Value Masks}
To further account for potential discrepancies between imputed values and actual observations, the missing value masks (Fig. \ref{Fig:Model_Architecture}(a)) are incorporated as an additional input to the model. We define the missing value masks for patient \(p\) as \(\tilde{X}_p\) with \([\tilde{X}_p]_{t,m} = 1\) if \([Y_p]_{t,m}\) is missing; otherwise \([\tilde{X}_p]_{t,m} = 0\). The masks will be  processed by an independent time-aware self-attention encoder in parallel with the imputed sequences as shown in Fig. \ref{Fig:Model_Architecture}(b). This missingness-aware design enables the network to capture the relationship between non-missing and missing values. As a result, the network is capable of mitigating the effect of potential errors during the MGP imputation process, thereby enhancing the overall predictive performance of the model. 

\subsubsection{Multi-branching Outputs}
The self-attention encoder serves as a powerful feature extractor from irregular longitudinal EHRs for downstream classification tasks. The classifier network, typically with a fully connected layer, is added after the feature extractor to interpret the abstracted features and make the final prediction. The imbalanced data issue incurs a more pronounced and direct negative impact on the classifier than on the feature extractor network. This is due to the fact that the classifier network directly interacts with the labels and is heavily influenced by the majority class during the training process \cite{zhou2020bbn}. This leads to a bias towards the majority class, visibly affecting the model's performance. As such, careful design of the classifier network is needed to cope with the imbalanced data issue. 

We propose to incorporate the Multi-branching (MB) architecture \cite{wang2021multi} (see Fig.\ref{Fig:Model_Architecture} (c)) into the classifier network to tackle the imbalanced issue. Specifically, in the training phase, the self-attention encoder will be trained with the whole dataset, and each of the MB outputs in the classifier network will be trained with a balanced sub-dataset to mitigate the negative influence of imbalanced data. The original imputed dataset \(\hat{\mathcal{D}}\) consists of the majority class \(\hat{\mathcal{D}}_M\), and the minority class \(\hat{\mathcal{D}}_N\). We create \(N_b\) balanced sub-datasets by under-sampling \(\hat{\mathcal{D}}_M\) to form \(\mathcal{D}_i = \{\hat{\mathcal{D}}_N, \hat{\mathcal{D}}_M^i\}\), where the \(\hat{\mathcal{D}}_M = \hat{\mathcal{D}}_M^1 \cup \hat{\mathcal{D}}_M^2 \cup \cdots \cup \hat{\mathcal{D}}_M^{N_b}\). This under-sampling operation is also applied to the missing value masks.  Correspondingly, \(N_b\) output branches are created, each aligned with one of the balanced sub-datasets. Thus, each balanced sub-dataset serves as the training data for the respective branch in the output layer, and the self-attention encoder will be optimized by using all the \(N_b\) balanced subsets. In the end, the MB output layer will produce \(N_b\) predicted probabilities. The optimization process is guided by minimizing the cross-entropy loss function:
{\small
\begin{eqnarray}
\mathcal{L}(\vect{\omega}; \hat{\mathcal{D}}) &=& - \sum_{p=1}^{N} \sum_{i=1}^{N_b} I(p \in \mathcal{D}_i) 
\Big( l_p \log(\hat{P}^{(i)}( \hat{Y}_p, \tilde{X}_p;\vect{\omega})) \nonumber \\
&& + (1 - l_p) \log(1 - \hat{P}^{(i)}( \hat{Y}_p, \tilde{X}_p;\vect{\omega})) \Big)
\label{obj_func}
\end{eqnarray}}
where \(\vect{\omega}\) is the network parameters; \(I(\cdot)\) is an indicator function; {\small\(\hat{P}^{(i)}(\hat{Y}_p, \tilde{X}_p;\vect{\omega})\)} is the prediction by branch \(i\) given the input $\hat{Y}_p$ and corresponding missing value mask \(\tilde{X}_p\). The
final predicted probability is computed as the average of the \(N_b\) predictions: {\small\(\hat{P}_{MB}(\hat{Y}_p, \tilde{X}_p;\vect{\omega}) = \sum_{i = 1}^{N_b}\hat{P}^{(i)}(\hat{Y}_p, \tilde{X}_p;\vect{\omega})/N_b\)}, which is a more robust estimator compared to the single-branch (SB) counterpart as stated in Theorem I.

\noindent\textbf{Theorem I}:  If both MB and SB classifiers are sufficiently trained with the following assumptions:
   \begin{eqnarray}
   \bar{\hat{P}}^{(i)}(Y_p)=\bar{\hat{P}}_{SB}(Y_p), ~~~ (i\in\{1,\dots,N_b\}) \nonumber\\
       \mathbb{E}_{\vect{\omega}}[D_{KL}(\bar{\hat{P}}^{(i)}(Y_p)||\hat{P}^{(i)}(Y_p;\vect{\omega}))]=\nonumber\\
       \mathbb{E}_{\vect{\omega}}[D_{KL}(\bar{\hat{P}}_{SB}(Y_p)||\hat{P}_{SB}(Y_p;\vect{\omega}))],
  \end{eqnarray}
where $\hat{P}_{SB}(Y_p;\vect{\omega})$ is the predicted distribution by the SB classifier for input $Y_p$, $D_{KL}(\cdot)$ is the Kullback–Leibler divergence, {\small$\bar{\hat{P}}_{SB}(Y_p) = \arg\min_{P_*}\mathbb{E}_{\vect{\omega}}[D_{KL}(P_*(Y_p)||\hat{P}_{SB}(Y_p;\vect{\omega}))]$}, and {\small$\bar{\hat{P}}^{(i)}(Y_p) = \arg\min_{P_*}\mathbb{E}_{\vect{\omega}}[D_{KL}(P_*(Y_p)||\hat{P}^{(i)}(Y_p;\vect{\omega}))]$}, then the variance of the MB classifier is no larger than the variance of the SB classifier, i.e., $V_{MB}\leq V_{SB}$, where
{\small
 \begin{eqnarray}
     V_{MB}&=&\mathbb{E}_{Y_p\sim P_{data},\vect{w}}[D_{KL}(\bar{\hat{P}}_{MB}(Y_p)||\hat{P}_{MB}(Y_p;\vect{w}))] \\
     V_{SB}&=&\mathbb{E}_{Y_p\sim P_{data},\vect{w}}[D_{KL}(\bar{\hat{P}}_{SB}(Y_p)||\hat{P}_{SB}(Y_p;\vect{w}))]
 \end{eqnarray}}
Here, $P_{data}$ denotes the true data distribution and 
   \begin{eqnarray}
       \bar{\hat{P}}_{MB}(Y_p) &=& \arg\min_{P_*}\mathbb{E}_{\vect{\omega}}[D_{KL}(P_*(Y_p)||\hat{P}_{MB}(Y_p;\vect{\omega}))] \label{Eq: MB_avg_def}
   \end{eqnarray}

 Theorem I shows that the MB classifier is more robust than the SB classifier (see Appendix \ref{appendix_mb} for the theoretical proof).


\section{Experimental Design and Results}


We implement our MUSE-Net to investigate both a synthetic and real-world dataset for performance evaluation. The MUSE-Net is benchmarked with four models widely used in sequential data mining: LSTM \cite{hochreiter1997long}, gated recurrent unit (GRU) \cite{chung2014empirical}, Time-aware LSTM (T-LSTM) \cite{baytas2017patient},  and TCN \cite{bai2018empirical}. Additionally, we evaluate the prediction performance provided by different imputation methods, i.e., the proposed MGP+mask, MGP, single GP, GP+mask, the baseline mean imputation, and Mean+mask. We also evaluate the impact of the number of branches on the final prediction performance. The dataset is split into training (80\%), validation (10\%), and testing (10\%) sets. The model performance is evaluated on four key metrics: area under the receiver operating characteristic curve (AUROC), area under the precision-recall curve (AUPRC), recall, and the F1 score. AUROC is a measure of the model's ability to discriminate between classes. AUPRC is valuable for assessing performance in classifying imbalanced datasets, reflecting the balance between precision and recall. Recall assesses the model's ability to correctly identify positive instances, while the F1 score provides a comprehensive evaluation by accounting for both the model’s sensitivity to positive cases and its overall predictive reliability.

\begin{algorithm}[ht]
\caption{Synthetic Data Generation Process}
\begin{algorithmic}[1]
\footnotesize
\Require Number of time steps: \(n\_obs=200\); Number of observations sampled from each time series: \(sub\_time = 50\); Number of variables: \(n\_variables=10\); Number of samples to generate: \(n\_samples=5000\); Percentage of the majority samples: \(percent\_negative=90\%\)
\State Initialize \(\Omega_m\) for ARMA(\(p_m\),\(q_m\)), \(m\in \{1,2,3\}\), and apply feature engineering to generate remaining ARMA(\(p_m\),\(q_m\)) \(m\in \{4,5,\cdots, 2 \times n\_variables\}\)
\State Generate a label list, \(labels\_list\), including \(n\_samples\) binary labels (0 or 1) based on $percent\_negative$
\State Generate empty 3-D tensor, \(all\_samples\), to store the generated data
\For {$j \gets 1$ \textbf{to} \(n\_samples\)}
    \State Initialize an empty temporal list: \(variables \gets [\quad]\)
    \For{$m \gets 1$ to $n\_variables$}
        \If{$label\_list[j]$ is $0$ (negative instance)} 
            \State Generate \(n\_variables\) time series using \(\text{ARMA}(p_m,q_m) \)
        \Else  
            \State Generate \(n\_variables\) time series using  \(\text{ARMA}(p_{m+n\_variables},q_{m+n\_variables}) \)
        \EndIf
        \State Add generated time series to $variables$
    \EndFor
    \State Combine $variables$ into a matrix with $n\_variables$ columns and add it to \(all\_samples\)

\EndFor
\State Initialize an empty tensor: \(final\_dataset\)
\For{$i \gets 1$ \textbf{to}  $n\_samples$}
    \State \(subtime\_indices \gets \) Maintains the temporal order and randomly selects \(sub\_time\) unique time steps
    \State \(sub\_ts\_i \gets all\_samples\)[\(i\), \(subtime\_indices\), :]
    \State Add generated \(sub\_ts\_i\) to \(final\_dataset\)
    \EndFor

\State \Return \(final\_dataset\)
\end{algorithmic}
\end{algorithm}

\subsection{Experimental Results in Synthetic Case Study}

Algorithm 1 shows the process to generate synthetic data from the autoregressive moving average (ARMA) model \cite{adhikari2013introductory}. Specifically, we first generate three baseline time series variables by ARMA(\(p_m\),\(q_m\)) models with moving average of order \(q_m\) and autoregressive of order \(p_m\):  \(v_{m,t} = \phi_{m,1}v_{m,t-1} + \cdots + \phi_{m,p_m}v_{m,t-p_m} + \psi_{m,1}\epsilon_{m,t-1} + \cdots + \psi_{m,q_m}\epsilon_{m, t-q_m} + \epsilon_{m, t}\), where \(\{\epsilon_{m, t}\}\) are uncorrelated random variables for variable \(m\); \(\Omega_m =\{\phi_{m,1}, \cdots, \phi_{m,p_m}, \psi_{m,1}, \cdots, \psi_{m,q_m}\}\) is the set of coefficients for variable \(m\). We initialize  \(\Omega_m\) randomly from a standard normal distribution and randomly select \(p_m\),\(q_m\) from set \(\{1,2,3\}\). We generate the remaining seven variables through feature engineering applied to the three baseline time series, ensuring correlations exist among the variables and differences exist between different classes. Detailed information on the feature engineering process and the simulation parameters is available in the Appendix \ref{appendix_arma}. The generated synthetic dataset is a 3-D tensor containing 5,000 samples with 90\% from the majority class. Each sample has 10 variables measured across 50 time steps with irregular intervals. Notably, each variable is characterized by a distinct rate of missing values, varying randomly between 30\% and 60\%, mimicking real-world data complexities. 

    \begin{figure*}[!ht]
	\begin{center}
		\includegraphics[width=5.5in]{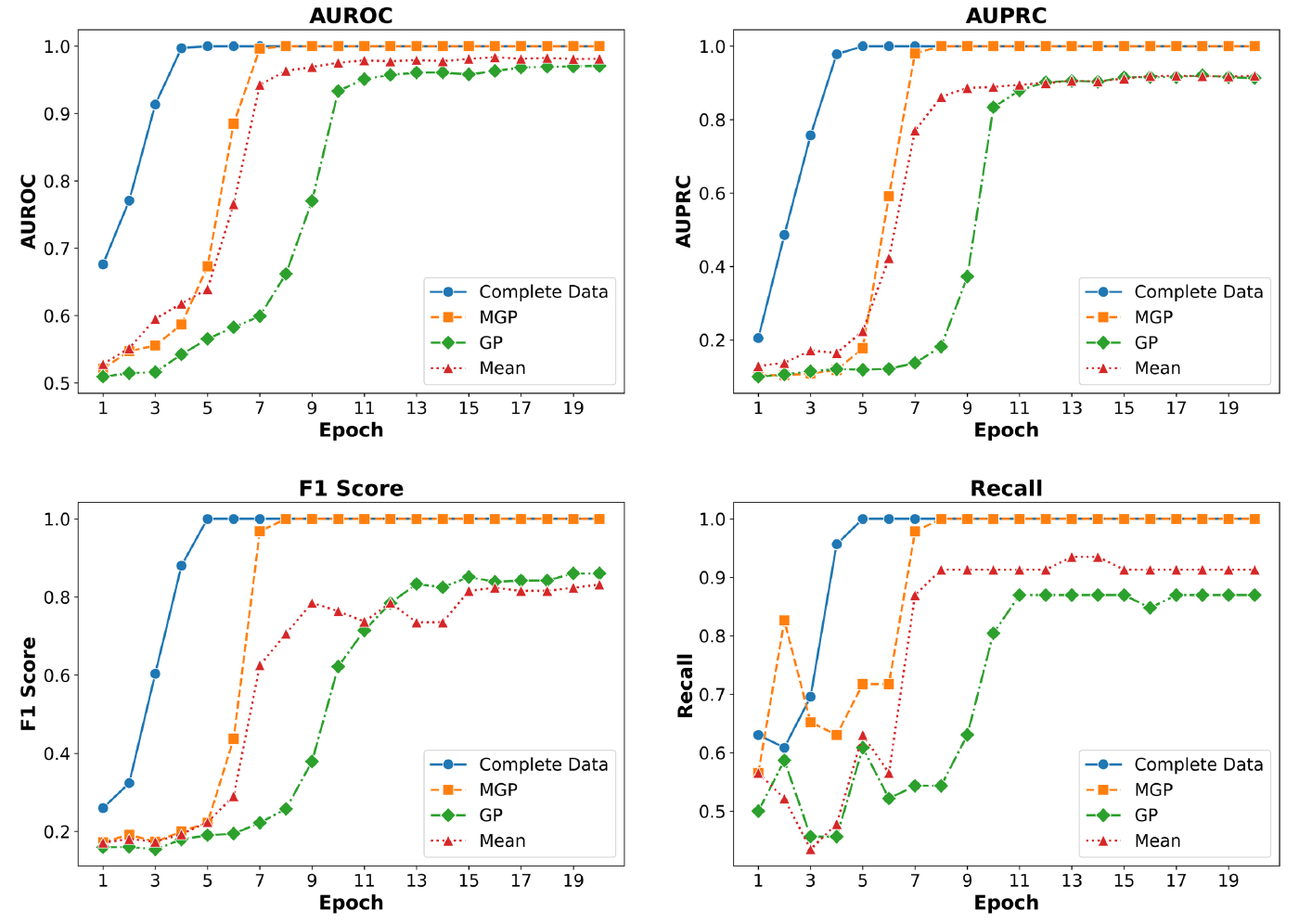}
		\caption{Evaluation scores across epochs for MUSE-Net-9 with different imputation methods on simulated validation data.}
		\label{Fig:Sim_imputation}
	\end{center}    
  
\end{figure*}
   
 \begin{figure*}[!ht]
	\begin{center}
		\includegraphics[width=5.5in]{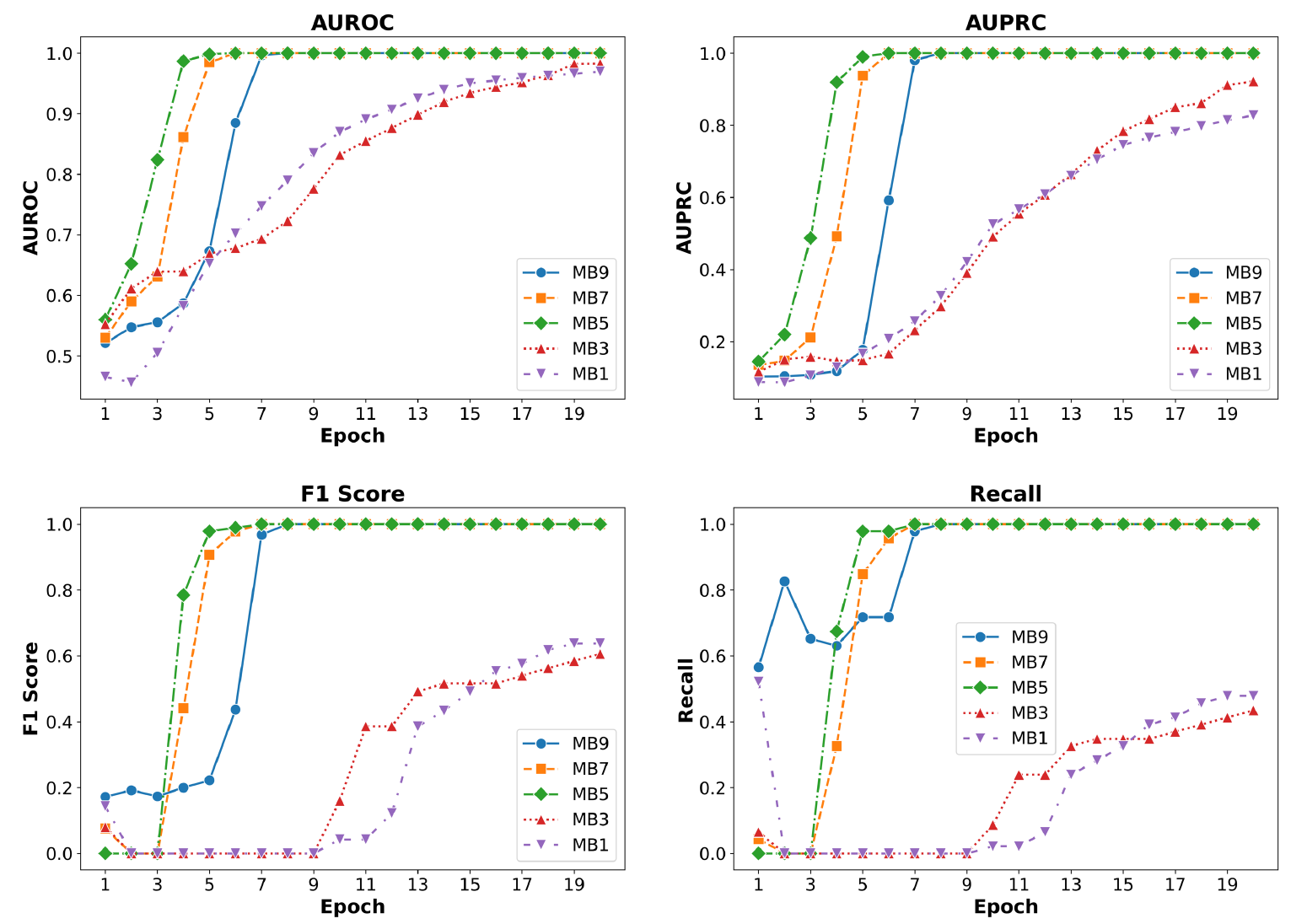}
		\caption{Evaluation metrics scores across epochs for MUSE-Net with varying MB outputs on MGP-imputed validation data.}
		\label{Fig:Sim_MB_Nb}
	\end{center}    
 
\end{figure*}

Fig. \ref{Fig:Sim_imputation} shows the AUROC, AUPRC, F1 score, and Recall on validation dataset over training epochs for MUSE-Net-9 (i.e., with 9 branching outputs) using different imputation methods with missing value masks on the validation set of our simulated data. The 'complete' scenario represents the case where MUSE-Net is trained on synthetic data without any missing values. Among all methods, MUSE-Net-9+MGP demonstrates the best performance across all four metrics, achieving higher AUROC, AUPRC, F1 score, and Recall, along with the fastest convergence. In contrast, the MUSE-Net-9+GP and MUSE-Net-9+mean methods exhibit slower convergence and lower overall performance, highlighting their limitations in handling missing values effectively.

Fig. \ref{Fig:Sim_MB_Nb} presents the AUROC, AUPRC, F1 score, and Recall across training epochs for MUSE-Net+MGP with different numbers of MB outputs on the simulated validation set. 
Models with a higher number of MB outputs (MB5, MB7, and MB9) demonstrate faster convergence and superior overall performance. This improvement can be attributed to the enhanced class balance within each branching output. When only one MB output is used (MB1), the class imbalance is severe (approximately 9:1), making it difficult for the model to effectively learn from the minority class. In contrast, increasing the number of MB outputs to 9 results in a more balanced class ratio (1:1) for each branch, facilitating better learning from both classes.

The F1 score and Recall curves further emphasize the differences in learning dynamics. Models with more MB outputs exhibit rapid improvements in these metrics, while those with fewer MB outputs (MB1 and MB3) show delayed growth and lower overall performance, particularly in the early training stages. This suggests that models with fewer MB outputs struggle to effectively capture minority class patterns, leading to slower convergence and reduced recall. Overall, increasing the number of MB outputs enhances both convergence speed and classification performance, reinforcing the benefits of a more balanced learning framework.

  \begin{table*}[!ht]
\centering
\caption{Architectures of MUSE-Net and other benchmarks for DR dataset.}
\label{model_info}
\small
\begin{tabular}{c|c|c|c|c}
\hline
 & \makecell[c]{Number of \\ model parameters} & \makecell[c]{Number of \\ layers (blocks)} & \makecell[c]{Number of outputs \\ in MB layer} & Optimizer \\ \hline
MUSE-Net & 4,010 & \multirow{4}{*}{2} & \multirow{4}{*}{10} & \multirow{4}{*}{\makecell[c]{Adam with \\ decoupled weight decay \\ (AdamW) \cite{loshchilov2017decoupled}}} \\ \cline{1-2}
GRU & 4,333 &  &  \\ \cline{1-2}
LSTM & 4,686 &  &  \\ \cline{1-2}
T-LSTM & 4,660 &  &  \\ \cline{1-2}
TCN & 4,396 &  &  \\ \hline
\end{tabular}
\end{table*}

\subsection{Experimental Results in Real-world Case Study}

We further conduct a case study using the Diabetic Retinopathy (DR) dataset, obtained from the 2018 Cerner Health Facts data warehouse \cite{wang2024multi,chen2022prediction}, to evaluate our MUSE-Net. The variables selected for this study include 21 routine blood tests, 5 comorbidity indicators, 3 demographic variables, and the duration of diabetes. More details on variable selection can be referred to \cite{chen2022prediction,wang2021derivation}. Notably, the 21 blood tests are all subject to missing values, and the final dataset consists of records from 23,245 diabetic patients, with a minority of 8.9\% diagnosed with DR. We use label ``1" and ``0" to indicate that a patient diagnosed with and without DR, respectively.

\begin{figure*}[!ht]
	\begin{center}
		\includegraphics[width=5.5in]{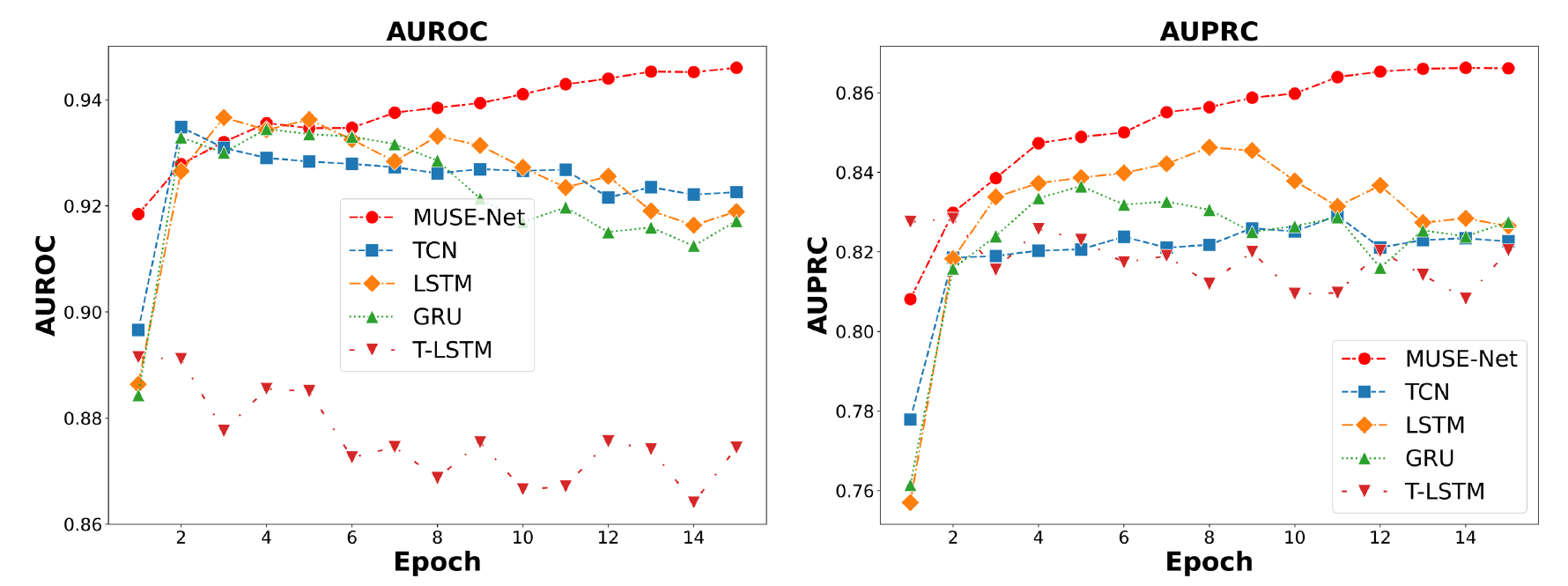}
		\caption{AUROC and AUPRC across 15 epochs for our MUSE-Net and other benchmarks on the DR validation set. }
		\label{Fig:DR_comp}
	\end{center}    
\end{figure*}

Fig. \ref{Fig:DR_comp} shows the performance of MUSE-Net compared to other benchmark models on the DR validation set. Note that to ensure a fair comparison, all models are designed to have similar sizes, with the number of model parameters approximately 4,000, as shown in Table \ref{model_info}. Furthermore, all models utilize the MGP imputation with missing value masks, 10 MB outputs, and Adam with decoupled weight decay optimizer \cite{loshchilov2017decoupled}. According to Fig. \ref{Fig:DR_comp}, our MUSE-Net consistently outperforms other benchmarks, achieving the highest AUROC and AUPRC on the validation set. This trend is also evident in the results for the test set summarized in Table \ref{tab:Model_scores_test}, where MUSE-Net maintains the AUROC of 0.949 and the AUPRC of 0.883, which dominates those yielded by TCN, LSTM, T-LSTM, and GRU models. Specifically, the MUSE-Net has an improvement over TCN by 2.3\% in AUROC and 5.3\% in AUPRC, and an improvement over LSTM by 3.3\% in AUROC and 7.1\% in AUPRC,  an improvement over T-LSTM by 2.4\% in AUROC and 6.8\%, and an improvement over GRU 1.9\% in AUROC and 4.6\% in AUPRC. This performance demonstrates that MUSE-Net is more effective at capturing the critical patterns inherent in irregular longitudinal DR data. 

{\small
\begin{table}
\caption{AUROC, AUPRC, Recall and F1 scores for MUSE-Net and other benchmarks on the DR test set. }
\centering
\begin{tabular}{lccccc}
\toprule
& MUSE-Net & TCN & LSTM & T-LSTM & GRU \\
\midrule
AUROC & 0.952 & 0.935 & 0.919 & 0.932 & 0.932 \\
AUPRC & 0.886 & 0.848 & 0.825 & 0.844 & 0.844 \\
F1    & 0.750 & 0.704 & 0.580 & 0.671 & 0.671 \\
Recall& 0.850 & 0.809 & 0.836 & 0.827 & 0.827 \\
\bottomrule

\label{tab:Model_scores_test}
\end{tabular}
\end{table}
}

Table \ref{tab:imputation_comp} presents the performance comparison in AUROC and AUPRC between our MUSE-Net-10 with MGP imputation and other imputation methods on the DR test set. It is worth noting that the MUSE-Net-10 with MGP+masks outperforms other methods, yielding an AUROC of 0.952 and an AUPRC of 0.886,  which represents an improvement over GP+masks by 2.0\% in AUROC and 7.1\% in AUPRC, and an improvement over mean+masks by 2.5\% in AUROC and 1.5\% in AUPRC. Moreover, consistent performance improvement is achieved across all imputation methods when missing value masks are applied. Specifically, MGP+masks achieves an improvement over the non-mask MGP by 5.7\% in AUROC and 10.6\% in AUPRC. 
The results show that the missing value masks enable the model to effectively account for the missingness patterns and mitigate potential discrepancies between imputed and actual values that may arise during the imputation process.

{\small
\begin{table}
\caption{AUROC and AUPRC comparison of MUSE-Net-10 with MGP and other imputation methods on the DR test set.}
\centering
\begin{tabular}{lcccccc}
\toprule
\makecell[c]{Model: \\ MUSE-Net-10} & \multicolumn{2}{c}{MGP} & \multicolumn{2}{c}{GP} & \multicolumn{2}{c}{Mean} \\
\cmidrule(r){2-3} \cmidrule(lr){4-5} \cmidrule(l){6-7}
& Mask & w/o & Mask & w/o & Mask & w/o \\
\midrule
AUROC & 0.952 & 0.901 & 0.933 & 0.889 & 0.929 & 0.885 \\
AUPRC & 0.886 & 0.801 & 0.827 & 0.791 & 0.876 & 0.789 \\
\bottomrule
\end{tabular}
\label{tab:imputation_comp}
\end{table}
}

Table \ref{tab:MB_scores_test} shows the AUROC and AUPRC scores for MUSE-Net with MGP imputation across various numbers of MB outputs in the DR test set. MB 1 corresponds to a model with a single output handling the original dataset with an imbalance ratio of approximately 1:10, and MB 10 represents a model with 10 outputs, each of which is trained with one of the 10 balanced sub-datasets with a ratio of approximately 1:1. This table suggests an overall increasing trend in both AUROC and AUPRC when the number of MB outputs increases. The highest AUROC/AUPRC scores of 0.952/0.886 are achieved with 10 MB outputs, demonstrating the enhanced predictive capability of our MUSE-Net in a balanced training environment. The performance gradually declines with a decrease in the number of MB outputs, reaching its lowest AUROC and AUPRC of 0.940 and 0.864 at MB 1. 

\begin{table}
\centering
\caption{AUROC and AUPRC scores for MUSE-Net+MGP with different MB numbers in the DR test set.}

\begin{tabular}{ccccccc}
\hline
& MB 10 & MB 8 & MB 6 & MB 4 & MB 2 & MB 1 \\
\hline
AUROC & 0.952 & 0.947 & 0.946 & 0.947 & 0.945 & 0.940\\
AUPRC & 0.886 & 0.876 & 0.877 & 0.886 & 0.869 & 0.864 \\

\hline
\end{tabular}

\label{tab:MB_scores_test}
\end{table}

\subsection{Intepretability Analysis of MUSE-Net in DR Prediction}

We further analyze the attention weights learned by the interpretable multi-head attention mechanism to interpret the model’s decision-making process for DR prediction. 

\subsubsection{Averaged Attention Map Across Layers}

\begin{figure}[!ht]
	\begin{center}
		\includegraphics[width=3.5in]{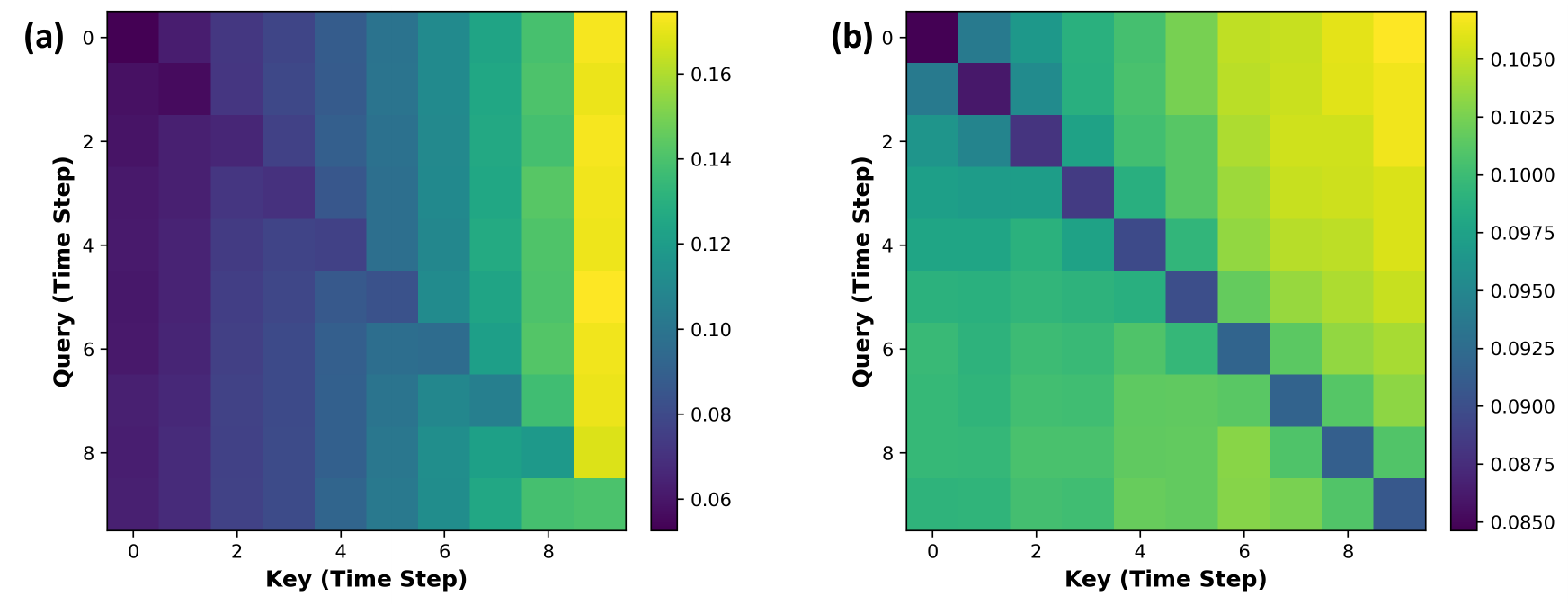}
		\caption{The averaged attention maps over all test samples for the first and second layers of the MUSE-Net: (a) the first layer; (b) the second layer}
		\label{Fig:Attention_Map}
        \vspace{-0.5cm}
	\end{center}    

\end{figure}
Fig. \ref{Fig:Attention_Map} presents the averaged attention maps over all test samples for the first and second layers of the MUSE-Net. Each heatmap represents the attention weight distribution between query (y-axis) and key time steps (x-axis), where higher values indicate stronger attention. These maps help clinicians interpret how the model makes predictions by identifying which time steps play a dominant role in decision-making. In the first layer (Fig. \ref{Fig:Attention_Map}(a)), the attention distribution is skewed toward the later time, with the highest attention at the final time step. This suggests that the model places greater importance on more recent time steps when making predictions. Earlier time steps receive lower attention, indicating that they contribute less to the decision-making in the initial processing layer. Clinically, this aligns with the understanding that recent variable changes carry the most relevant information for assessing DR progression. 

In the second layer (Fig. \ref{Fig:Attention_Map}(b)), attention weights are more evenly distributed. This indicates that, after the first layer refines feature representations, the second layer integrates a more comprehensive view of temporal dependencies, balancing both recent and earlier observations. From a clinical perspective, this suggests that while initial assessments focus on recent trends, a deeper analysis incorporates long-term history to ensure a comprehensive decision-making process. This hierarchical mechanism mirrors how clinicians evaluate patient histories: first prioritizing the most recent indicators, then considering long-term trends for a more thorough assessment.

\subsubsection{Attention Patterns in DR and Non-DR}
\begin{figure}[!ht]
	\begin{center}
		\includegraphics[width=3.5in]{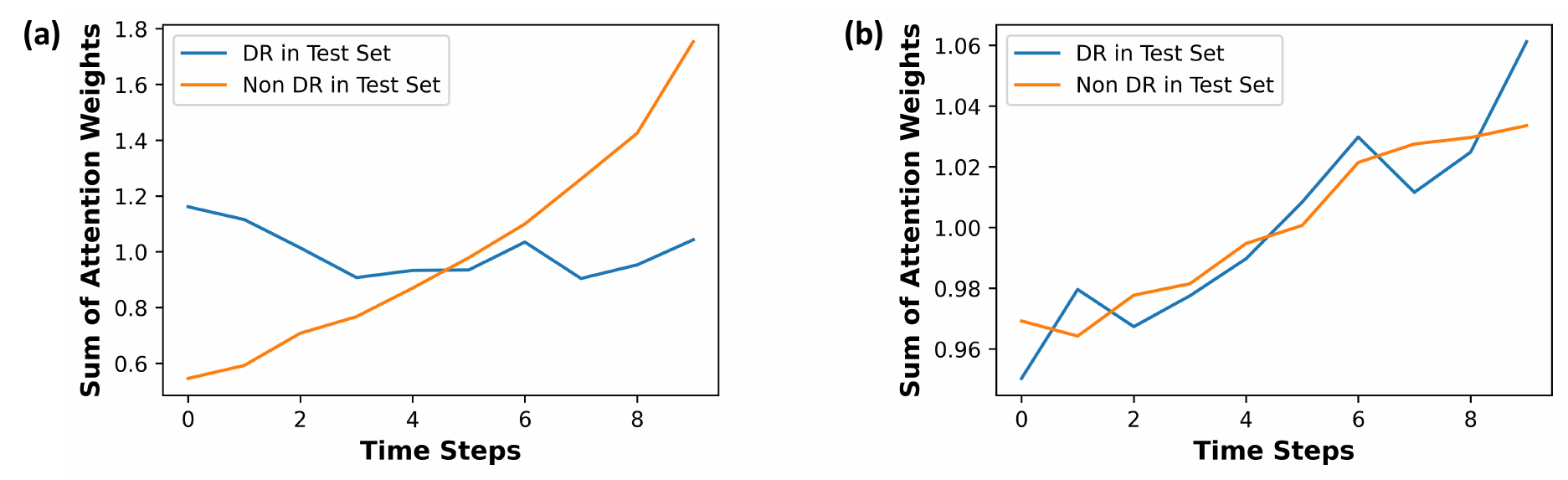}
		\caption{The column-wise sum of attention weights across all time steps in the test dataset: (a) the first layer; (b) the second layer}
		\label{Fig:Sum_Attention}
	\end{center}    
    \vspace{-0.5cm}
\end{figure}

{
Fig. \ref{Fig:Sum_Attention} shows the column-wise sum of attention weights across all time steps for DR and Non-DR patients, providing insight into how much total attention is allocated to each time step for different groups in the test dataset. Fig. \ref{Fig:Sum_Attention}(a) shows the sum of attention weights in the first layer, revealing a clear difference: DR samples exhibit a more evenly distributed attention pattern across time steps. This suggests that the model considers multiple time steps when identifying DR, which is clinically relevant because DR progression is a gradual process influenced by long-term trends of many risk factors. Non-DR samples, on the other hand, show a steady increase in attention allocation over time, meaning that recent observations are more predictive of non-DR. This aligns with clinical practice where the absence of warning signs in recent medical history reduces the disease likelihood. This divergence in attention distribution provides critical insights: for DR patients, historical data remains important in assessing disease progression, whereas for non-DR patients, absence of recent warning signs is a stronger predictor of continued healthy status. 

Fig. \ref{Fig:Sum_Attention}(b) presents the summed attention weights from the second layer. Here, both DR and non-DR samples follow an increasing trend, but a key observation is that the scale of the y-axis is smaller than in the first layer, indicating that attention is more evenly distributed. This suggests that each time step retains its importance, ensuring that no single time step dominates the decision-making process for both DR and Non-DR patients. These findings demonstrate that MUSE-Net not only achieves strong predictive performance but also offers clinically interpretable insights by mimicking how human experts evaluate temporal patient data in DR diagnosis.}

\section{Conclusions}

In this paper, we introduce a novel framework: \(\mathbf{M}\)issingness-aware m\(\mathbf{U}\)lti-branching \(\mathbf{S}\)elf-Attention \(\mathbf{E}\)ncoder (MUSE-Net) to model irregular longitudinal EHRs with missingness and imbalanced data issues. First, multi-task Gaussian processes (MGPs) are leveraged for missing value imputation in irregularly sampled longitudinal signals. Second, we propose a time-aware self-attention encoder augmented with a missing value mask and multi-branching architectures to classify irregular longitudinal EHRs. Furthermore, an
interpretable multi-head attention mechanism is added to the model to highlight critical time points in disease
prediction, offering transparency in decision-making and enabling clinicians to trace model outputs
back to influential time points. Finally, we evaluate our proposed framework using both simulation data and real-world EHRs. Experimental results show that our MUSE-Net significantly outperforms existing approaches that are widely used to investigate longitudinal signals. More importantly, this framework can be broadly applicable to model complex longitudinal data with issues of multivariate irregularly spaced time series, incompleteness, and imbalanced class distributions.




\section{Appendix}

\subsection{Parameter Specifications in Synthetic Data Generation }
\label{appendix_arma}

Tables \ref{tab:Parameters_ARMA} and \ref{tab:feature_process_ARMA} provide the parameter specification in ARMA models and the feature engineering process.

\begin{table}[!ht]
\small
\caption{Parameter specifications of the three baseline ARMA models}
\vspace{+0.2cm}
\label{tab:Parameters_ARMA}
\centering
\begin{tabular}{l|c|c}
\hline
& \(p_m, q_m\) & \(\Omega_m\) \\
\hline
\(\text{ARMA}_1\), \(m = 1\) & (2, 2) & \{-0.75, 0.25, 0.65, 0.35\} \\
\(\text{ARMA}_2\), \(m = 2\)  & (1, 1) & \{-0.8, 0.5\} \\
\(\text{ARMA}_3\), \(m = 3\)  & (3, 3) & \{-0.65, 0.45, -0.2, 0.70, 0.45, 0.25\} \\
\hline
\end{tabular}
\end{table}

\begin{table}[!ht]
\footnotesize
\caption{Feature engineering process to generate time series for each variable.}
\label{tab:feature_process_ARMA}
\centering
\begin{tabularx}{\columnwidth}{l X X}
\hline
\(m\) & Neg. samples (label = 0) & Pos. samples (label = 1) \\
\hline
1 & \(\text{ARMA}_1\) & \(\text{ARMA}_{11} = \text{ARMA}_1\) \\
2 & \(\text{ARMA}_2\) & \(\text{ARMA}_{12} = \text{ARMA}_2\) \\
3 & \(\text{ARMA}_3\) & \(\text{ARMA}_{13} = \text{ARMA}_3\) \\
4 & \(\text{ARMA}_4 = 0.8 \times \text{ARMA}_1\) & \(\text{ARMA}_{14} = 1.1 \times \text{ARMA}_{11}\) \\
5 & \(\text{ARMA}_5 = \text{ARMA}_1 + \text{ARMA}_2\) & \(\text{ARMA}_{15} = \text{ARMA}_{11} + \text{ARMA}_{12}\) \\
6 & \(\text{ARMA}_6 = \text{ARMA}_3 + \text{ARMA}_4\) & \(\text{ARMA}_{16} = \text{ARMA}_{13}+ \text{ARMA}_{14}\) \\
7 & \(\text{ARMA}_7 = \text{ARMA}_1 + \text{ARMA}_2 + \text{ARMA}_3\) & \(\text{ARMA}_7 = \text{ARMA}_{11} + \text{ARMA}_{12} + \text{ARMA}_{13}\) \\
8 & \(\text{ARMA}_8 =\text{ARMA}_1 \times \text{ARMA}_2 + \text{ARMA}_3 \times \text{ARMA}_4\) & \(\text{ARMA}_{18} =\text{ARMA}_{11} \times \text{ARMA}_{12} + \text{ARMA}_{13} \times \text{ARMA}_{14}\) \\
9 & \(\text{ARMA}_9 = \text{ARMA}_3 + \text{ARMA}_4 + \text{ARMA}_1 \times \text{ARMA}_2\) & \(\text{ARMA}_{19} = 1.1 \times \text{ARMA}_{13} + 1.1 \times \text{ARMA}_{14} + \text{ARMA}_{11} \times \text{ARMA}_{12}\)  \\
10 & \(\text{ARMA}_{10} = -\text{ARMA}_5 + \text{ARMA}_9\) & \(\text{ARMA}_{20} = -\text{ARMA}_{15} + \text{ARMA}_{19}\) \\
\hline
\end{tabularx}
\end{table}


\subsection{Robust Classifier from Multi-branching Network Models}
\label{appendix_mb}

   This appendix shows the proof of Theorem I.

   \textbf{Definition I} (Bregman Divergence): If $F:\mathcal{X} \rightarrow \mathbb{R}$ is a convex differentiable function, the Bregman Divergence based on $F$ is a function $D_F: \mathcal{X}\times \mathcal{X} \rightarrow \mathbb{R}_+$, defined as
       \begin{eqnarray}
       	 D_F[x_1||x_2]\equiv F(x_1)-F(x_2)-\langle \nabla F(x_2), x_1-x_2\rangle, \nonumber \\ x_1, x_2 \in \mathcal{X} 
       \end{eqnarray}  

   \textbf{Lemma I} (Generalized bias-variance decomposition \cite{pfau2013generalized}): Let $F:\mathcal{X} \rightarrow \mathbb{R}$ be a convex differentiable function, $g(Y_p)$ is the true function, and $f(Y_p;\vect{\omega})$ is the prediction, where $\vect{\omega}$ is the model parameter, the generalized bias-variance decomposition based on the Bregman divergence $D_F$ is 
     \begin{eqnarray}
     	\mathbb{E}_{Y_p,\vect{\omega}} [D_F(g(Y_p)||f(Y_p;\vect{\omega}))] = \nonumber \\
        \mathbb{E}_{Y_p,\vect{\omega}} [D_F(g(Y_p)||\bar{f}(Y_p))] + \mathbb{E}_{Y_p,\vect{\omega}} [D_F(\bar{f}(Y_p)||f(Y_p;\vect{\omega}))] 
     	\label{gbv}
     \end{eqnarray}
   where $\bar{f}(Y_p) = \arg\min_z\mathbb{E}_{\vect{\omega}}[D_F(z||f(Y_p;\vect{\omega}))]$.
   
\begin{proof} (Lemma I)
From the definition of $\bar{f}(Y_p)$, we have:
\begin{eqnarray}
0 &=& \nabla_z \mathbb{E}_{\vect{\omega}}[D_F(z||f(Y_p;\vect{\omega}))] \Big|_{z=\bar{f}(Y_p)} \nonumber\\
  &=& \nabla_z \mathbb{E}_{\vect{\omega}} \Big[F(z) - F(f(Y_p;\vect{\omega})) \nonumber\\
  && - \langle \nabla F(f(Y_p;\vect{\omega})), z - f(Y_p;\vect{\omega}) \rangle \Big] \Big|_{z=\bar{f}(Y_p)} \nonumber \\
  &=& \Big[ \nabla F(z) - \nabla_z \mathbb{E}_{\vect{\omega}} 
  \langle \nabla F(f(Y_p;\vect{\omega})), z \rangle \Big] \Big|_{z=\bar{f}(Y_p)} \nonumber\\
  &=& \Big[ \nabla F(z) - \mathbb{E}_{\vect{\omega}}[ \nabla F(f(Y_p;\vect{\omega})) ] \Big] \Big|_{z=\bar{f}(Y_p)} \nonumber\\
  &\Rightarrow& \nabla F(\bar{f}(Y_p)) = \mathbb{E}_{\vect{\omega}}[ \nabla F(f(Y_p;\vect{\omega})) ].
\end{eqnarray}

Then, the right-hand side of Eq. (\ref{gbv}) can be rearranged as
\begin{eqnarray}
   &&\mathbb{E}_{Y_p,\vect{\omega}} [D_F(g(Y_p)||\bar{f}(Y_p))] 
   + \mathbb{E}_{Y_p,\vect{\omega}} [D_F(\bar{f}(Y_p)||f(Y_p;\vect{\omega}))] \nonumber\\
   &=& \mathbb{E}_{Y_p,\vect{\omega}} \Big[F(g(Y_p)) - F(\bar{f}(Y_p)) \nonumber\\
   &&\quad - \langle \nabla F(\bar{f}(Y_p)), g(Y_p)-\bar{f}(Y_p) \rangle \Big] \nonumber\\
   && + \mathbb{E}_{Y_p,\vect{\omega}} \Big[F(\bar{f}(Y_p)) - F(f(Y_p;\vect{\omega})) \nonumber\\
   &&\quad - \langle \nabla F(f(Y_p;\vect{\omega})), \bar{f}(Y_p) - f(Y_p;\vect{\omega}) \rangle \Big] \nonumber\\
   &=& \mathbb{E}_{Y_p,\vect{\omega}} \Big[F(g(Y_p))-F(f(Y_p;\vect{\omega})) \nonumber\\
   &&\quad - \langle \nabla F(\bar{f}(Y_p)), g(Y_p)-\bar{f}(Y_p) \rangle \nonumber\\
   &&\quad - \langle \nabla F(f(Y_p;\vect{\omega})), \bar{f}(Y_p)-f(Y_p;\vect{\omega}) \rangle \Big] \nonumber\\
   &=& \mathbb{E}_{Y_p,\vect{\omega}} \Big[F(g(Y_p))-F(f(Y_p;\vect{\omega})) \nonumber\\
   &&\quad - \langle \nabla F(f(Y_p;\vect{\omega})),g(Y_p)-f(Y_p;\vect{\omega}) \rangle \Big] \nonumber\\
   &=& \mathbb{E}_{Y_p,\vect{\omega}} [D_F(g(Y_p)||f(Y_p;\vect{\omega}))].
\end{eqnarray}

   \end{proof}
   
   The bias-variance decomposition for classification analysis with cross-entropy loss is provided in Theorem II:

   \textbf{Theorem II} (Bias-variance decomposition for classification tasks): We denote the true class distribution for input $Y_p$ as $P(Y_p)$ and the predicted distribution given by the classification model as $\hat{P}(Y_p,\vect{w})$, where $\vect{w}$ is the model parameter set. We assume there are $J$ classes in total. Then, according to the generalized decomposition for Bregman divergence, the bias-variance decomposition for classification tasks is given as
\begin{eqnarray}
   \mathbb{E}_{Y_p,\vect{\omega}} [D_{KL}(P(\vect{x})||\hat{P}(Y_p,\vect{w}))] 
   = \mathbb{E}_{Y_p,\vect{\omega}} [D_{KL}(P(Y_p)||\bar{\hat{P}}(Y_p))] \nonumber\\
   + \mathbb{E}_{Y_p,\vect{\omega}} [D_{KL}(\bar{\hat{P}}(Y_p)||\hat{P}(Y_p;\vect{\omega}))].
   \label{gbv2}
\end{eqnarray}
   where $\bar{\hat{P}}(Y_p) = \arg\min_{P_*}\mathbb{E}_{\vect{\omega}}[D_{KL}(P_*(Y_p)||\hat{P}(Y_p;\vect{\omega}))]$ and $D_{KL}(P||\hat{P})=\sum_{j=1}^Jp_j\log\frac{p_j}{\hat{p}_j}$ is the Kullback-Leibler (KL) divergence. The first term in Eq. (\ref{gbv2}) is the squared bias, and the second term captures the variance of the classifier model.

    \begin{proof} (Theorem II)
        Note that minimizing KL divergence is equivalent to minimizing cross-entropy loss in classification tasks. The cross-entropy loss is defined as $
     	H(P||\hat{P})=-\sum_{j=1}^Jp_j\log \hat{p}_j = D_{KL}(P||\hat{P})-\sum_j(p_j\log p_j)$,
        where $\sum_j(p_j\log p_j)$ is often considered as a constant. Hence, minimizing $H(P||\hat{P})$ is equivalent to minimizing $D_{KL}(P||\hat{P})$. Additionally, KL divergence is a special case of the Bregman divergence. Specifically, if we define the function in Definition I as $F(P)=\sum_{i}p_i\log p_i$, then
    \begin{eqnarray}
    	&&D_F(P||\hat{P}) \nonumber\\
        &=& \sum_i(p_i\log p_i)-\sum_i(\hat{p}_i\log \hat{p}_i)  -\sum_i\langle \log \hat{p}_i+1, p_i-\hat{p}_i\rangle \nonumber\\
        &=&\sum_i p_i\log \frac{p_i}{\hat{p}_i} =D_{KL}(P||\hat{P})
    \end{eqnarray}
  
    As such, we can use Lemma I for the Bregman divergence to prove Theorem II by defining  $F(P)=\sum_{i}p_i\log p_i$, and replacing $g(Y_p)$ by the true distribution $P(Y_P)$ and replacing the prediction $f(Y_p;\vect{\omega})$ by $\hat{P}(Y_p;\vect{\omega})$.
   \end{proof}

   Finally, the following shows the proof of Theorem I:
   \begin{proof} (Theorem I)
       From Theorem II, the variances of SB and MB classifiers are (omitting subscripts $Y_p$ for notation convenience)
       {
     \begin{eqnarray}
         V_{SB} =\mathbb{E}_{\vect{\omega}}[D_{KL}(\bar{\hat{P}}_{SB}||\hat{P}_{SB}(\vect{\omega}))], \\ \nonumber V_{MB} =\mathbb{E}_{\vect{\omega}}[D_{KL}(\bar{\hat{P}}_{MB}||\hat{P}_{MB}(\vect{\omega}))]
     \end{eqnarray}
     }
   where $\hat{P}_{SB}(\vect{\omega})$ and $\hat{P}_{MB}(\vect{\omega})$ are the predictions provided by the SB and MB classifiers. In the MB model, the predicted probability for class $j$ is $\hat{p}_{MB,j}(\vect{\omega})=\frac{1}{N_b}\sum_{i=1}^{N_b}\hat{p}_j^{(i)}(\vect{\omega})$, where $\hat{p}_j^{(i)}(\vect{\omega})$ is the predicted probability for class $j$ by branch $i$. Given the assumptions in Theorem I, i.e., $ \bar{\hat{P}}^{(i)}=\bar{\hat{P}}_{SB}$ and $\mathbb{E}_{\vect{\omega}}[D_{KL}(\bar{\hat{P}}^{(i)}||\hat{P}^{(i)}(\vect{\omega}))]=\mathbb{E}_{\vect{\omega}}[D_{KL}(\bar{\hat{P}}_{SB}||\hat{P}_{SB}(\vect{\omega}))]$ ($i\in\{1,\dots,N_b\}$) if all the branches and the SB model are sufficiently trained, 
  we have
    {\footnotesize
    \begin{eqnarray}
        V_{SB}&=&\frac{1}{N_b}\sum_{i=1}^{N_b}\mathbb{E}_{\vect{\omega}}[D_{KL}(\bar{\hat{P}}_{SB}||\hat{P}_{SB}(\vect{\omega}))]\nonumber\\
        &=&\frac{1}{N_b}\sum_{i=1}^{N_b}\mathbb{E}_{\vect{\omega}}[D_{KL}(\bar{\hat{P}}^{(i)}||\hat{P}^{(i)}(\vect{\omega}))]\nonumber\\
        &=& \frac{1}{N_b}\sum_{i=1}^{N_b}\mathbb{E}_{\vect{\omega}}\left[\sum_{j=1}^J\bar{\hat{p}}_{j}^{(i)}\log \frac{\bar{\hat{p}}_{j}^{(i)}}{\hat{p}_j^{(i)}(\vect{\omega})}\right] \nonumber\\
        &=&\mathbb{E}_{\vect{\omega}}\left[\sum_{j=1}^J\left(\bar{\hat{p}}_{SB,j}\log \bar{\hat{p}}_{SB,j}-\sum_{i=1}^{N_b}\bar{\hat{p}}_{SB,j}\frac{\log\hat{p}_j^{(i)}(\vect{\omega})}{N_b}\right)\right]\nonumber\\
        &=&\mathbb{E}_{\vect{\omega}}\left[\sum_{j=1}^J\left(\bar{\hat{p}}_{SB,j}\log \bar{\hat{p}}_{SB,j}-\bar{\hat{p}}_{SB,j}\log(\Pi_{i=1}^{N_b}\hat{p}_j^{(i)}(\vect{\omega}))^{\frac{1}{N_b}}\right)\right] \nonumber\\
        &\geq& \mathbb{E}_{\vect{\omega}}\left[\sum_{j=1}^J\left(\bar{\hat{p}}_{SB,j}\log \bar{\hat{p}}_{SB,j}-\bar{\hat{p}}_{SB,j}\log(\frac{1}{N_b}\sum_{i=1}^{N_b}\hat{p}_j^{(i)}(\vect{\omega}))\right)\right]\nonumber\\
        &=& \mathbb{E}_{\vect{\omega}}\left[\sum_{j=1}^J\left(\bar{\hat{p}}_{SB,j}\log \bar{\hat{p}}_{SB,j}-\bar{\hat{p}}_{SB,j}\log\hat{p}_{MB,j}(\vect{\omega})\right)\right]\nonumber\\
        &=&\mathbb{E}_{\vect{\omega}}[D_{KL}(\bar{\hat{P}}_{SB}||\hat{P}_{MB}(\vect{\omega}))]\geq \mathbb{E}_{\vect{\omega}}[D_{KL}(\bar{\hat{P}}_{MB}||\hat{P}_{MB}(\vect{\omega}))] \nonumber\\
        &=&V_{MB}
    \end{eqnarray}
    } 
     where the first inequality is true due to the fact that the geometric mean of nonnegative variables is always less than or equal to the arithmetic mean, and the second inequality is true due to the definition of $\bar{\hat{P}}_{MB}$ as given in Eq. (\ref{Eq: MB_avg_def}). As such, the MB classifier is more robust than the SB classifier, i.e., $V_{MB}\leq V_{SB}$, as stated in Theorem I. 
     \end{proof}

 \subsection{Algorithm for GPU Acceleration in MGP Imputation}
 \label{appendix mgp}

In our study, we utilize GPU acceleration for training MUSE-Net, which is a fundamental practice in modern deep learning. However, using GPUs to speed up MGP imputation remains relatively unexplored and requires further investigation. MGP-based imputation can be computationally intensive because it involves operations of large kernel matrices. 
Three key components contribute to the computational intensity: multiplication of the inverse kernel matrix with observed values \((\Sigma_p^o)^{-1} \vect{y}_p^o\), computation of log determinant \(\log |\Sigma_p^o|\) and trace \(\text{Tr} \left( (\Sigma_p^o)^{-1} \frac{d\Sigma_p^o}{d\vect{\Theta}} \right)\). In most existing MGP imputation techniques \cite{zhang2022multi, futoma2017learning, moor2019early, rosnati2021mgp}, calculation of the three quantities hinges on Cholesky decomposition of \((\Sigma_p^o)^{-1}\), which has cubic time complexity. Furthermore, the nonparallelizable nature of Cholesky decomposition makes it not suitable for GPU acceleration. To address this challenge, we adopt the blackbox matrix-matrix multiplication framework \cite{gardner2018gpytorch} to integrate the modified preconditioned conjugate gradient (mPCG) approach with GPU acceleration into the data imputation workflow.

Algorithm 2 summarizes the procedure for MGP imputation with mPCG. The inputs are initialized MGP hyperparameters \(\vect{\Theta}_0\), observed value \(\vect{y}_p^{o}\), covariance matrix \(\Sigma_p^o\), a matrix \(\Delta = [\vect{d}_0, \vect{d}_1, \cdots, \vect{d}_{t}] \in \mathbb{R}^{O_p \times (t+1)}\), where \(\vect{d}_0 = \vect{y}_p^o\), and \(\vect{d}_1, \vect{d}_2, \cdots, \vect{d}_t\) are \(i.i.d\) random vectors from a probability distribution with \(\mathbb{E}(\vect{d}_i) = 0\) and \(\mathbb{E}(\vect{d}_i \vect{d}_i^T) = \mathcal{P}\) ($i\in\{1,\dots,t\}$), \(\mathcal{P} = CC^T = (C_0C_0^T + E) \approx \Sigma_p^o\), where \(C_0C_0^T \approx \mathcal{K}_M \odot \mathcal{K}_{O_p}\) is a low-rank matrix approximation of \(\mathcal{K}_M \odot \mathcal{K}_{O_p}\) generated by pivotal Cholesky decomposition with $C_0\in\mathbb{R}^{O_p\times r_0}$ ($r_0\ll O_p$) \cite{harbrecht2012low}. \(\mathcal{P}^{-1}\) will then be employed as a preconditioner for \(\Sigma_p^o\) to enhance the efficiency of the mPCG \cite{gardner2018gpytorch}. The output of Algorithm 2 is the posterior mean for missing value \(\mu_{\vect{y}_p^{u}}\). Algorithm 2 consists of two main steps: the MGP training step with mPCG acceleration (from line 5 to 15) to generate optimal hyperparameters, \(\vect{\Theta}^{*}\), from lines 3 to 21, followed by the imputation step from line 22 to the end of the algorithm. 
  
In MGP training, the key is to compute \((\Sigma_p^o)^{-1}\Delta = [(\Sigma_p^o)^{-1}\vect{d}_0, (\Sigma_p^o)^{-1}\vect{d}_1, (\Sigma_p^o)^{-1}\vect{d}_2, \cdots, (\Sigma_p^o)^{-1}\vect{d}_t]\) in a parallel way in GPU using mPCG (see line 5 to 15). After \(k\) iterations, the output of mPCG, i.e., \(U_{k} \approx (\Sigma_p^o)^{-1}\Delta\) and \(\mathcal{T}_1, \mathcal{T}_2 , \cdots, \mathcal{T}_t \in \mathbb{R}^{k \times k}\), which are partial Lanczos tridiagonalizations \cite{golub2013matrix} of \(\Sigma_p^o\), can be used to compute \((\Sigma_p^o)^{-1} \vect{y}_p^o\), \(\log |\Sigma_p^o|\), and \(\text{Tr} \left( (\Sigma_p^o)^{-1} \frac{d\Sigma_p^o}{d\vect{\Theta}} \right)\) efficiently. Specifically, \((\Sigma_p^o)^{-1} \vect{y}_p^o\) can be obtained directly as  \((\Sigma_p^o)^{-1} \vect{y}_p^o \approx [U_{k}]_{:,1}\), where $[U_{k}]_{:,1}$ is the first column of matrix $U_{k}$. The computation of \(\text{Tr} \left( (\Sigma_p^o)^{-1} \frac{d\Sigma_p^o}{d\vect{\Theta}} \right)\) depends on stochastic trace estimation \cite{gardner2018gpytorch, ubaru2017fast}, which is a method used to efficiently approximate the trace of a large matrix. The basic idea is to use random vectors, i.e., \(\vect{d}_1, \vect{d}_2, \cdots, \vect{d}_t\), to probe the matrix as:
{\footnotesize
\begin{eqnarray}
\text{Tr} \left( (\Sigma_p^o)^{-1} \frac{d\Sigma_p^o}{d\vect{\Theta}} \right) &=& \text{Tr} \left( (\Sigma_p^o)^{-1} \frac{d\Sigma_p^o}{d\vect{\Theta}}\underset{\vect{d}_i \sim \mathcal{N}(0, \mathcal{P})}{\mathbb{E}}[\mathcal{P}^{-1}\vect{d}_i\vect{d}_i^T] \right) \nonumber\\
&=& \underset{\vect{d}_i \sim \mathcal{N}(0, \mathcal{P})}{\mathbb{E}}\left[\text{Tr} \left((\Sigma_p^o)^{-1} \frac{d\Sigma_p^o}{d\vect{\Theta}} \mathcal{P}^{-1}\vect{d}_i\vect{d}_i^T \right)\right] \nonumber\\
&=&\underset{\vect{d}_i \sim \mathcal{N}(0, \mathcal{P})}{\mathbb{E}}\left[\vect{d}_i^T (\Sigma_p^o)^{-1} \left(\frac{d\Sigma_p^o}{d\vect{\Theta}} \mathcal{P}^{-1}\vect{d}_i\right)\right] \nonumber\\
&\approx& \frac{1}{t} \sum_{i=1}^{t}\left(\vect{d}_i^T (\Sigma_p^o)^{-1}\right)\left(\frac{d\Sigma_p^o}{d\vect{\Theta}} \mathcal{P}^{-1}\vect{d}_i \right)
\label{trace}
\end{eqnarray}
}
where \(\vect{d}_i^T (\Sigma_p^o)^{-1}\approx ([U_{k}]_{:,i+1})^T\) are computed by mPCG, and $\mathcal{P}^{-1}$ can be calculated using the Matrix Inversion Lemma, i.e., $\mathcal{P}^{-1}=(C_0C_0^T + E)^{-1}=E^{-1}-E^{-1}(I+C_0^TE^{-1}C_0)^{-1}C_0^TE^{-1}$, which avoids the inversion computation of the $O_p\times O_p$ matrix using the inversion of a smaller $r_0\times r_0$ matrix. To estimate \(\log |\Sigma_p^o|\), we integrate stochastic trace estimation with partial Lanczos tridiagonalizations \cite{golub2013matrix, saad2003iterative}, i.e., \(\mathcal{T}_1, \mathcal{T}_2, \cdots, \mathcal{T}_t\), from mPCG. The key idea is that instead of calculating the full tridiagonalization, i.e., \((C^{-1})^{T}\Sigma_p^oC^{-1} = H^T\mathcal{T}H\) where \(H\) is orthonormal, we run \(k\) iterations (\(k \ll O_p\)) of this algorithm \(t\) times so that we obtain \(t\) partial Lanczos tridiagonalizations decompositions \(\mathcal{T}_i\) (see line 13 and 14 in Algorithm 1). Specifically, the log determinant \(\log |\Sigma_p^o|\) can be calculated by:
{\small
\begin{eqnarray}
&&\log |\Sigma_p^o| = \log |\Sigma_p^o C^{-1}\mathcal{P}(C^{-1})^{T}| \nonumber\\
&=&  \log |(C^{-1})^{T}\Sigma_p^oC^{-1}| + \log |\mathcal{P}| \\
&&\log |(C^{-1})^{T}\Sigma_p^oC^{-1}| = \text{Tr}(\log \mathcal{T}) \nonumber\\
&=& \underset{\vect{d}_i \sim \mathcal{N}(0, \mathcal{P})}{\mathbb{E}}[(\vect{d}_i^T(C^{-1})^{T})(\log \mathcal{T})(C^{-1}\vect{d}_i)] \nonumber \\
&=& \underset{\vect{d}_i \sim \mathcal{N}(0, \mathcal{P})}{\mathbb{E}}[(\vect{d}_i^T(C^{-1})^{T})H_i^T(\log \mathcal{T}_i)H_i(C^{-1}\vect{d}_i)] \nonumber\\
&\approx& \frac{1}{t}\sum_{i=1}^{t} \vect{e}_1^T(\log T_i)\vect{e}_1
\label{logdet}
\end{eqnarray}
}
where $\log |\mathcal{P}|$ can be computed using the Matrix Determinant Lemma as $\log |\mathcal{P}|=\log |C_0C_0^T + E|=\log |C_0^TE^{-1}C_0 + I|+\log |E|$, which converts the determinant calculation of $O_p\times O_p$ matrix into the determinant calculation of $r_0\times r_0$ matrix; \(\mathcal{T}_i \in \mathbb{R}^{k \times k}\) are partial Lanczos tridiagonal matrices, which can approximate the eigenvalue of \((C^{-1})^{T}\Sigma_p^oC^{-1}\) and can be obtained directly from mPCG (see line 13 and 14) \cite{golub2013matrix, saad2003iterative}; \(H_i\) is orthonormal matrix generated by \(C^{-1}\vect{d}_i\) so that \(H_i(C^{-1}\vect{d}_i) = \vect{e}_1\), where \(\vect{e}_1\) is the first column of identity matrix. Many existing studies \cite{gardner2018gpytorch, wang2019exact} have shown that by setting an iteration number \(k \ll O_p\) and utilizing GPU to compute matrix operations in parallel, the mPCG algorithm can approximate exact solutions of MGP but with much faster speed compared to Cholesky decomposition. 

The calculated \((\Sigma_p^o)^{-1} \vect{y}_p^o\), \(\log |\Sigma_p^o|\), and \(\text{Tr} \left( (\Sigma_p^o)^{-1} \frac{d\Sigma_p^o}{d\vect{\Theta}} \right)\) are used for computing \(\mathcal{L}(\vect{\Theta})\) and its derivative \(d\mathcal{L}/d\vect{\Theta}\) (see lines 17 and 18). These values will be further utilized to estimate the MGP hyperparameters through a gradient-based optimization algorithm (i.e., ADAM Optimizer \cite{kingma2014adam}), denoted as \(\vect{\Theta}^{*}\), which is subsequently utilized to compute the posterior mean, completing data imputation as shown in line 22.

\begin{algorithm}[!ht]
\small
\caption{Multi-task Gaussian Process (MGP) imputation with modified preconditioned conjugate gradients (mPCG) acceleration }
\begin{algorithmic}[1]
\State \textbf{Input:} Initialization of \(\vect{\Theta}_0\); \( \vect{y}_p^o\); \(\Sigma_p^o\); \( \Delta = [\vect{d}_0, \vect{d}_1, \cdots, \vect{d}_{t}]\); \( \mathcal{P}^{-1} = (CC^T)^{-1} = (C_0C_0^T + E)^{-1} \), \(C_0C_0^T\) generated by decomposing \(\mathcal{K}_M\odot \mathcal{K}_{O_p}\) using pivot Cholesky decomposition; \(l \gets 0\)
\State \textbf{Output:} The posterior mean for missing values \( \mu_{\vect{y}_p^u} \)
\While{\( \text{stopping criterion not met} \)} \Comment{Start training step}
    \State \( U_0 \gets 0 \); \( R_0 \gets \Sigma_p^o U_0 - \Delta\); \( Z_0 \gets \mathcal{P}^{-1}(R_0) \); \( S_0 \gets -Z_0 \); \( \mathcal{T}_1, \ldots, \mathcal{T}_t \gets 0 \) 
    \For{\( j = 0 \) \textbf{to} \( k-1 \)} \Comment{Execute mPCG }
        \State \( \vect{\alpha}_j \gets (R_{j} \odot Z_{j})^T \vect{1}/(S_{j} \odot (\Sigma_p^o S_{j}))\vect{1} \)
        \State \( U_{j+1} \gets U_{j} +  S_{j}\text{diag}(\vect{\alpha}_j) \)
        \State \( R_{j+1} \gets R_{j} +  \Sigma_p^o S_{j}\text{diag}(\vect{\alpha}_j) \)
        \If{\( \forall i \in{1,2, \cdots, t} \quad \|[R_{j+1}]_{:,i}\|_2 < \text{tolerance} \)} \textbf{yield} \( U_{j+1} \) \(\quad\) \textbf{break}
        \EndIf
        \State \( Z_{j+1} \gets \mathcal{P}^{-1}(R_{j+1}) \)
        \State \( \vect{\beta}_{j} \gets (Z_{j+1} \odot Z_{j+1})^T \vect{1} /(Z_{j} \odot Z_{j})^{T} \vect{1}\)
        \State \( S_{j+1} \gets - Z_{j+1} +  S_{j}\text{diag}(\vect{\beta}_j) \)
        \State \(\forall i \in{1,2, \cdots, t}\) \quad \( [\mathcal{T}_i]_{j+1,j+1} \gets \frac{1}{[\vect{\alpha}_{j}]_i} + \frac{[\vect{\beta}_{j-1}]_i}{[\vect{\alpha}_{j-1}]_i} \) \text{if} \(j > 0\), \text{otherwise} \( [\mathcal{T}_i]_{j+1,j+1} \gets 1/[\vect{\alpha}_{j}]_i \)
        \State \(\forall i \in{1,2, \cdots, t}, j>0\) \quad \( [\mathcal{T}_i]_{j+1,j}, [\mathcal{T}_i]_{j,j + 1} \gets  \frac{\sqrt{[\vect{\beta}_{j}]_i}}{[\vect{\alpha}_{j}]_i}\) 
    \EndFor
    \State \textbf{yield} \( U_{k}, \mathcal{T}_1, \ldots, \mathcal{T}_t \) \Comment{Used for computation of \(\mathcal{L}(\vect{\Theta}_l)\) and \(\frac{d\mathcal{L}}{d\vect{\Theta}_l}\)}
    \State Compute: \((\Sigma_p^o)^{-1} \vect{y}_p^o \gets [U_{k}]_{:,1}, ~\log |\Sigma_p^o| \gets \sum_{i=1}^{t} \vect{e}_1^T(\log T_i)\vect{e}_1\) + \(\log |\mathcal{P}|\), \(\text{Tr} \left( (\Sigma_p^o)^{-1} \frac{d\Sigma_p^o}{d\vect{\Theta}_{l}} \right) \gets \frac{1}{t} \sum_{i=1}^{t}([U_{k}]_{:,i+1})^T\left(\frac{d\Sigma_p^o}{d\vect{\Theta}_{l}} \mathcal{P}^{-1}\vect{d}_i \right)\)

    \State \(\mathcal{L}(\vect{\Theta}_{l}) \gets \log |\Sigma_p^o| + (\vect{y}_p^o)^T (\Sigma_p^o)^{-1} \vect{y}_p^o + \frac{O_p}{2}\log(2\pi)\)
    \State \(\frac{d\mathcal{L}}{d\vect{\Theta}_{l} } \gets (\vect{y}_p^o)^T (\Sigma_p^o)^{-1} \frac{d\Sigma_p^o}{d\vect{\Theta}_{l}} (\Sigma_p^o)^{-1} \vect{y}_p^o + \text{Tr} \left( (\Sigma_p^o)^{-1} \frac{d\Sigma_p^o}{d\vect{\Theta}_{l}} \right)\)
    \State \(\vect{\Theta}_{l+1} \gets ADAMOptimizer\left(\mathcal{L}(\vect{\Theta}_{l}), \frac{d\mathcal{L}}{d\vect{\Theta}_{l} }\right)\) \Comment{Update \(\vect{\Theta}_{l+1}\) by ADAM optimizer}
    \State \(l \gets l + 1\)
\EndWhile
\\
\textbf{yield} \(\vect{\Theta}^{*} = \{\text{Vect}(B)^{*}, \text{Diag}(E)^{*}, \theta^{*} \}\) \Comment{Complete training step}
\State \(\mu_{\vect{y}_p^u} \gets (\mathcal{K}(\text{Vect}(B)^{*})_{M_{*}M} \odot \mathcal{K}(\theta^{*})_{U_pO_p})(\Sigma_p^o(\vect{\Theta}^{*}))^{-1} \vect{y}_p^o\) \Comment{Complete imputation step}
\State \Return \( \mu_{\vect{y}_p^u}\)
   
\label{MGP-mPCG}
\end{algorithmic}
\end{algorithm}

\bibliographystyle{IEEEtran}

\bibliography{ref}


\end{document}